\documentclass[default,iicol]{sn-jnl}

\makeatletter
\newlength{\continueindent}
\renewenvironment{algorithmic}[1][0]%
   {%
   \edef\ALG@numberfreq{#1}%
   \def\@currentlabel{\theALG@line}%
   \setcounter{ALG@line}{0}%
   \setcounter{ALG@rem}{0}%
   \let\\\algbreak%
   \expandafter\edef\csname ALG@currentblock@\theALG@nested\endcsname{0}%
   \expandafter\let\csname ALG@currentlifetime@\theALG@nested\endcsname\relax%
   \begin{list}%
      {\ALG@step}%
      {%
      \rightmargin\z@%
      \itemsep\z@ \itemindent\z@ \listparindent2em%
      \partopsep\z@ \parskip\z@ \parsep\z@%
      \labelsep 0.5em \topsep 0.2em
      \ifthenelse{\equal{#1}{0}}%
         {\labelwidth 0.5em}%
         {\labelwidth 1.2em}%
       \leftmargin\labelwidth \addtolength{\leftmargin}{\labelsep}
      \ALG@tlm\z@%
      }%
      \parshape 2 \leftmargin \linewidth \continueindent \dimexpr\linewidth-\continueindent\relax
   \setcounter{ALG@nested}{0}%
   \ALG@beginalgorithmic%
   }%
   {
   \ALG@closeloops%
   \expandafter\ifnum\csname ALG@currentblock@\theALG@nested\endcsname=0\relax%
   \else%
      \PackageError{algorithmicx}{Some blocks are not closed!!!}{}%
   \fi%
   \ALG@endalgorithmic%
   \end{list}%
   }%
\makeatother

\usepackage{hyperref}
\usepackage{amsmath}

\jyear{2023}%

\theoremstyle{thmstyleone}%
\theoremstyle{thmstyletwo}%
\newtheorem{remark}{Remark}%

\theoremstyle{thmstylethree}%

\begin{document}
\section*{Copyright notice}
\thispagestyle{empty}
This work has been submitted for possible publication. Copyright may be transferred without notice, after which this version may no longer be accessible.

\newpage

\title[Article Title]{Sustainable Cloud Services for Verbal Interaction with Embodied Agents}


\author*[1]{\fnm{Lucrezia} \sur{Grassi}}\email{lucrezia.grassi@edu.unige.it}

\author[1]{\fnm{Carmine Tommaso} \sur{Recchiuto}}\email{carmine.recchiuto@dibris.unige.it}

\author[1]{\fnm{Antonio} \sur{Sgorbissa}}\email{antonio.sgorbissa@unige.it}

\affil[1]{\orgdiv{Department of Computer Science, Bioengineering, Robotics and Systems Engineering}, \orgname{University of Genoa}, \orgaddress{\street{Via All'Opera Pia 13}, \postcode{16145}, \city{Genoa}, \country{Italy}}}


\abstract{This article presents the design and the implementation of a cloud system for knowledge-based autonomous interaction devised for Social Robots and other conversational agents. The system is particularly convenient for low-cost robots and devices: it can be used as a stand-alone dialogue system or as an integration to provide ``background" dialogue capabilities to any preexisting Natural Language Processing ability that the robot may already have as part of its basic skills. By connecting to the cloud, developers are provided with a sustainable solution to manage verbal interaction through a network connection, with about 3,000 topics of conversation ready for “chit-chatting" and a library of pre-cooked plans that only needs to be grounded into the robot's physical capabilities. The system is structured as a set of REST API endpoints so that it can be easily expanded by adding new APIs to improve the capabilities of the clients connected to the cloud. Another key feature of the system is that it has been designed to make the development of its clients straightforward: in this way, multiple robots and devices can be easily endowed with the capability of autonomously interacting with the user, understanding when to perform specific actions, and exploiting all the information provided by cloud services. The article outlines and discusses the results of the experiments performed to assess the system's performance in terms of response time, paving the way for its use both for research and market solutions. Links to repositories with clients for ROS and popular robots such as Pepper and NAO are available on request.}


\maketitle

\section{Introduction}
\label{sec:introduction}
\begin{figure}
    \centering
    \includegraphics[width=0.6\linewidth]{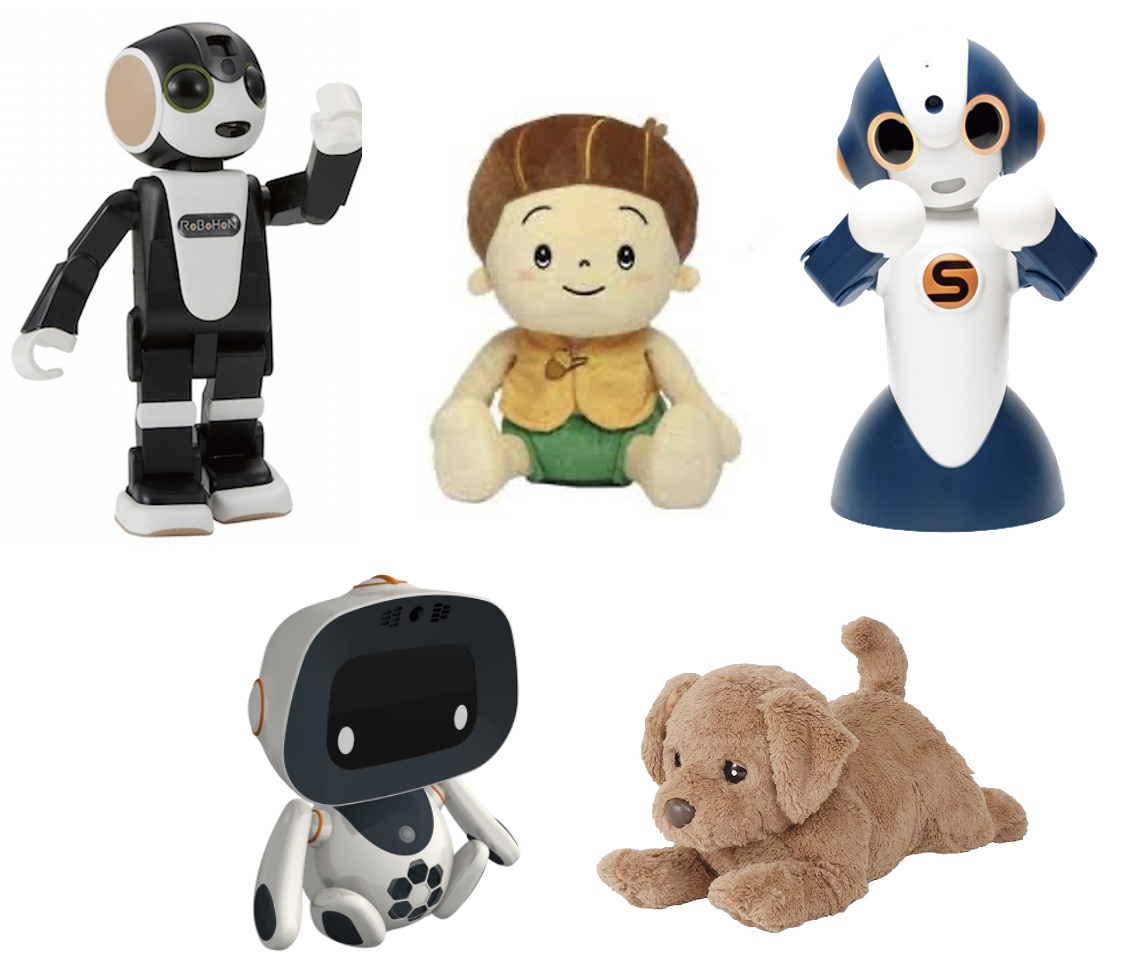}
    \caption{RoBoHoN, Kabo-chan, SOTA, UNIBO, and Dacky.}
    \label{fig:asian_robots}
\end{figure}

\begin{figure}
    \centering
    \includegraphics[width=0.6\linewidth]{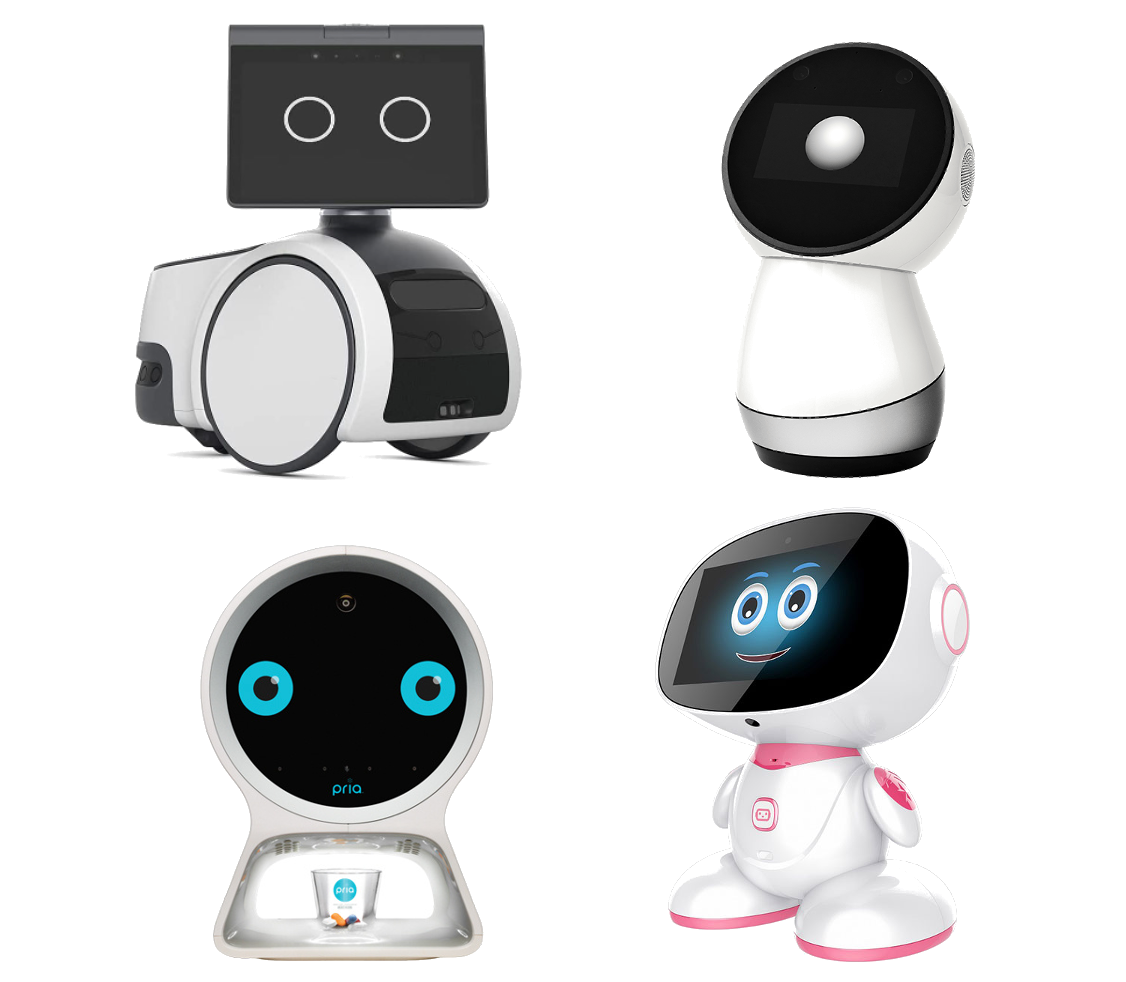}
    \caption[]{Amazon Astro and Echo Dot, Jibo, Pillo, and Misa\footnotemark.}
    \label{fig:western_robots}
\end{figure}

In recent years, it has become increasingly common to exploit cloud technologies to improve the efficiency of intelligent systems and devices. In the robotics field, this practice is defined as \textit{cloud robotics}, i.e., remote computing resources to enable greater memory, computational power, collective learning, and interconnectivity for robotics applications \cite{wan2016}.
Cloud-based solutions are particularly appealing considering the plethora of robots hitting the market daily. Most of these robots are often low-cost with limited capabilities in terms of sensors, actuating devices, and onboard computing power. Up to a decade ago, it was unlikely that a family had in their house a device smarter than a cleaning, mopping, or lawn-mowing robot. 
However, following the success of home assistants such as Google Home or Alexa, market reports estimate that we can expect to be soon ``invaded" by low-cost next-generation assistants and table-robots for health care, companionship, entertainment and education\footnote{For instance: \url{https://www.marketsandmarkets.com/Market-Reports/educational-robot-market-28174634.html} and \url{https://www.marketsandmarkets.com/Market-Reports/service-robotics-market-681.html}}. This process started a few years ago in Asia, with robots such as RoBoHoN, Kabo-chan, SOTA, UNIBO, and Dacky (Figure \ref{fig:asian_robots}), followed by Europe and the USA with robots such as Amazon Astro, Jibo, Pillo, and Misa (Figure \ref{fig:western_robots}) -- some of which already disappeared from the market. The process is expected to be further boosted as a consequence of the Covid-19 pandemic \cite{wan2020}.

Each of the aforementioned robots has its own abilities of perception and action, and the ability to interpret specific user commands and requests through either voice commands (e.g., ``Hey Google...", ``Alexa...") or a dedicated smartphone application (e.g., iRobot Home App). However, depending on the considered scenario, onboard capabilities might reveal insufficient. Think about assisting older people in care homes or younger people that cannot move from their bed due to a spinal cord injury, two scenarios that we recently explored in clinical trials with robots. Conversation, playing videos or music, reminding of daily events and activities, etc., could play a role in reducing loneliness, improving mood, and keeping people active \cite{papadopoulos2020, Papadopoulos2021}. However, even a robot like SoftBank's Pepper, despite its 17,000 Euros cost, has very limited capabilities when it comes out of the box, and its onboard processing power allows programmers to implement only the simplest behaviours.
In all these situations, the outsourcing of some robots' capabilities to third-party cloud-based solutions allows the devices to perform tasks that may be too heavy (and then expensive) for local processing, therefore revealing to be a low-cost option to improve performance. 

In this general scenario, two examples of activities that may be conveniently outsourced are:

\begin{enumerate}[(i)]
    \item Dialogue: the ability to autonomously converse with the users, covering numerous topics, always keeping into account users' preferences;
    \item Activity planning: the generation of complex plans “grounded" on the robot's sensorimotor capabilities;
\end{enumerate}

Based on these premises, the main research question we pose is the following:
\begin{quote}
Can a cloud system be designed to support sustainable, knowledge-grounded autonomous interaction between humans and embodied artificial agents while ensuring acceptable response times over a network, thus enabling social robots to perform these abilities?
\end{quote}


Please notice that we use the term “sustainable” with three meanings: 
\begin{enumerate}
    \item First, we need solutions apt to deal with the unprecedented growth of the social robotics market, providing acceptable performance and cost.
    The cloud should be structured in a way that can be easily expanded by adding new services to improve the capabilities of the robots and devices connected to it. Still, this responsibility should be on the shoulders of the developers and maintainers of the system and its services, not on the companies developing clients running on robots and devices. 
    \item Second, by writing an appropriate client application connecting to the cloud, the system should be easily usable by any robot and device with Internet connectivity, able to acquire an input through a keyboard or a microphone, and provide an output through a screen or a speaker (e.g.,
    robots, computers, smartphones, smartwatches, etc.), even very simple and low cost ones. 
    \item Third, and differently from other research or market solutions, the cloud should provide a complete solution that other researchers and companies can use without requiring additional programming or customization. 
    Customization should be possible but optional:  
    the system should come with a vast portfolio of diverse conversation topics ready for “chit-chatting" and a library of pre-cooked plans to be grounded into the robot's physical capabilities, and
    social robot developers should require no additional work other than connecting to the cloud to manage conversations or planning.  
\end{enumerate}

To address the research question above, the main contributions of this work are:

\begin{itemize}
    \item a cloud system  (CAIR, Cloud-based Autonomous Interaction with Robots) specifically designed for human-robot interaction and with the explicit objective of being sustainable along all the above dimensions, which presently comes with about $3,000$ conversation topics and plans suitable for the physical capabilities of SoftBank's Pepper and NAO;
    \item the experimental evaluation of the system to analyze its performance in terms of response time when multiple clients connect to the cloud, a fundamental measurement to validate the feasibility of the approach for human-robot interaction.
\end{itemize}


Finally, it is worth anticipating that the ability of the system to understand the user's intention and act accordingly (Section \ref{sec:plan}) and naturally converse with the user with mixed-initiative dialogues (Section \ref{sec:dialogue}) is based on a framework for cultural knowledge representation that relies on an OWL 2 Ontology \cite{guarino1998} paralleled by a framework for probabilistic reasoning. The framework is designed to take into account the possible cultural differences between different users in a non-stereotyped way, both concerning conversation and action \cite{recchiuto2020, bruno2019, grassi2021socrob}. Currently, the system takes into account four cultures (i.e., English, Italian, Japanese, and Indian) and it is available in three languages (i.e., English, Italian and Japanese).
A mechanism to expand in run-time the cultural knowledge base with new concepts and conversation topics raised by the users was introduced in \cite{grassi2022} and is currently being integrated as a feature of CAIR.

\begin{remark}
The focus of the present article is not on Natural Language Processing solutions but rather on technical solutions to manage the conversation between an embodied agent and a person by addressing the specific needs raised by a complex, goal-oriented, mixed-initiative dialogue.
\end{remark}
\begin{remark}
Similarly, the focus is not on end-user evaluation of the quality of the interaction (e.g., through questionnaires) but on the technical feasibility of implementing a cloud service that can be sustainable for real-time interaction with multiple heterogeneous devices. Concerning user evaluation, please note that the solutions for an engaging human-robot interaction described in this article have already been evaluated in a 7-month trial from March 2019 to September 2019 with older residents of care homes in the UK and Japan \cite{Papadopoulos2021}. More than 30 older residents verbally interacted with the system for 18 hours, divided into six 3-hour sessions, about various culturally-relevant topics, including their childhood, family, the place where they live, and many others. Additional trials have been performed with 300 middle school students (from December 2022 to February 2023) and people with spinal cord injuries at the Santa Corona hospital in Italy (ongoing). 
\end{remark}

The rest of the article is organized as follows. Section \ref{sec:SOTA} presents an overview of previous works on cloud robotics in different areas. Section \ref{sec:sys-arch} describes the architecture of the system, focusing on the design and implementation of the cloud server, and provides a description of the main operations that a client should perform (two examples of already implemented clients are given). Section \ref{sec:experiments} describes the experiment carried on to evaluate the system's performance in terms of response speed. Section \ref{sec:results} presents the results and discusses the outcome of the experiments. Eventually, Section \ref{sec:conclusion} draws the conclusion.

\section{Related Work}
\label{sec:SOTA}
Cloud robotics is a field of robotics that exploits cloud technologies such as cloud storage and cloud computing. Cloud storage allows a client robot, computer, tablet, or smartphone to send and retrieve files online to and from a remote data server. Cloud computing is the delivery of computing resources over the Internet, which may include storage, processing power, databases, networking, analytics, artificial intelligence, and dedicated applications. The use of a cloud for other autonomous systems may provide several benefits such as increased computational power, access to high storage space, easier collective robot learning thanks to the sharing of data and control policies, access to parallel grid computing on demand for statistical analysis, motion planning, and knowledge crowdsourcing \cite{wan2016, kehoe2015, saha2018}.

For instance, \cite{wan2018} analyzes a cognitive industrial entity called context-aware cloud robotics (CACR) for advanced material handling. The decision-making mechanism for energy-efficient and cost-saving material handling is managed by a cloud scheduler that searches for the platforms with the smallest number of workpieces. A command is sent to the nearest robots to divert them to the required platform and fulfill the material handling request. In this way, the mechanism uses the least amount of robotics resources to meet the requirements of all the manipulation platforms.

Another industrial example of cloud robotics is cloud-connected vehicles such as Google’s self-driving cars \cite{waymo}. Autonomous cars use the network to access Google's database of maps, satellite, and environment models (such as Street View) and combine those data with streaming data from GPS, cameras, and 3D sensors. This allows them to monitor their position and compare it with past and current traffic patterns to avoid collisions. Each car can gather information about road signs, pavement markings, lane closures, and traffic conditions and send it to the cloud. Such data are then processed and used to improve the performance and safety of all cars.

To explore the benefits of cloud-based technologies for sensing, \cite{mohanarajah2015} presented a cloud-based collaborative visual Simultaneous Localization and Mapping (SLAM) system consisting of low-cost robots and remote Amazon servers, which allows real-time map estimation through parallel computing and implements a highly sophisticated system using low-cost components, thus lowering the technological threshold for further research. 

Concerning complex tasks that a single robot is not able to perform, cloud robotics is widely accepted as a promising approach to efficient robot cooperation \cite{chen2018}, e.g. when dealing with search and rescue in disaster management. The article argues that the main benefits are the ability to cooperate within the disaster site and the possibility of distributing the algorithms on the remote cloud. Among the critical challenges, there is quality of service issues, communication issues, safety issues, heterogeneity, and security issues.

Other examples of cloud-based architectures in the industrial and service robotics domains are discussed in \cite{liu2016, cardarelli2017, yan2017}.

Despite the numerous benefits, \cite{wan2016} points out several key issues and challenges in cloud robotics, such as communication issues due to network problems, latency issues, interoperability, and portability due to the variety of infrastructures, platforms, and APIs offered by cloud providers, and privacy and security of personal data uploaded on the cloud. All these issues are related to the concept of a `“sustainable" human-robot interaction, which becomes particularly relevant when targeting low-cost consumer robots with limited capabilities designed for social interaction with people.

\subsection{Cloud-Based Social Robotics}
In the Social Robotics domain, the outsourcing of services to third-party services in the cloud mostly concerns the aspects already mentioned in Section \ref{sec:introduction}, i.e., (i) dialogue \cite{yan2016, ouerhani2020} and (ii) activity planning \cite{lam2014, joo2019, singhal2017, zagradjanin2019}, (iii) collective learning \cite{kaan2018, chibani2013, sandygulova2013, bonaccorsi2016} and (iv) perception \cite{soyata2012, pawle2013, hossain2015, muhammad2015, dinuovo2019}. All these aspects are discussed in the following.

\subsubsection{Dialogue}
Social robots and conversational agents need to communicate in a way that feels natural to humans to  bond with them and provide an engaging interaction. Concerning dialogue, there are several cloud platforms providing tools to build, test, and deploy an embodied conversational agent such as Dialogflow\footnote{\url{https://dialogflow.cloud.google.com}}, IBM Watson Assistant\footnote{\url{https://www.ibm.com/cloud/watson-assistant}}, Azure Bot Service\footnote{\url{https://azure.microsoft.com/en-us/services/bot-services/}}, and Amazon Lex\footnote{\url{https://aws.amazon.com/lex/}}. However, as explained in \cite{yan2016}, despite the advantages provided by such services, building an embodied conversational agent that meets the requirements for social interaction is still challenging both concerning interaction capabilities and from the technical perspective. 

Regarding the interaction capabilities, the aforementioned services have many drawbacks: they cannot properly manage a mixed-initiative dialogue since they only reply to the user's requests; it is not easy to customize them to show good performance in addressing many different topics; they are not easy to integrate with the actions that a robot can perform; they are not “grounded” with the data acquired by their sensors; they do not have a system for knowledge representation \cite{mavridis}; they are mostly available in English only. Even more importantly, the responsibility of creating “good” conversation systems is left to developers who decide to use these tools. This task can be costly, and developers may not be well-prepared to accomplish it, also considering that building an embodied agent with not-only-conversational capabilities will require the technical integration of the agent with multiple third-party services and additional issues linked to extensibility, scalability, and maintenance. 

To solve some of the technical issues related to the development of embodied agents, serverless computing \cite{mcgrath2017} has recently emerged as a way of creating backend applications. The cloud service provider of serverless applications automatically provisions, scales, and manages the infrastructure required to run the code, enabling developers to build applications faster. 
Major cloud vendors such as Google, Amazon, Microsoft, and IBM have created serverless versions of their frameworks, whose life-cycle costs are typically lower than costs for dedicated infrastructure as their vendors do not charge for idle time. Still, in most cases, the conversational agents built using the services provided by these cloud platforms are meant to be used in Q\&A scenarios such as providing customer support, booking tickets, and ordering food. See, for example, the serverless conversational agents that interact with various commodity services publicly available on the Internet, such as weather service, discussed in \cite{yan2016}. 
This type of conversational agent is designed to recognize the user's intent and respond to their specific requests based on its intended purpose. However, it does not engage in a complex, goal-oriented, mixed-initiative dialogue aimed at achieving human-robot interaction, which is the requirement for our scenario. 

Special attention should be devoted to Rospeex \cite{sugiura205}, a cloud platform designed especially for the robotics community providing high-quality multilingual speech recognition and synthesis engines accessible as ROS (Robot Operating System) modules. Other than providing such services, the aim of the platform is to collect log data that can be shared with the robotics community. However, Rospeex does not provide a ready-to-use solution for mixed-initiative dialogue about different topics: once again, it assumes that developers, typically with no experience in “playwriting" and no sensitivity on how to talk to people (possibly to frail people), write code for dialogue-related functions, including language understanding, dialogue management, and response generation. 

Now suppose that, despite their limitations, a researcher decides to use one of the aforementioned systems that indeed provide interesting Natural Language Processing functionalities for interpreting and replying to user commands, e.g., DialogFlow. In that case, the aforementioned systems might easily coexist with CAIR. With its $3,000$ conversation topics ready for a culturally-sensitive “chit-chatting” and its library of pre-cooked plans, the latter focuses on dialogue management in a broader sense providing a backbone for dialogue and planning capabilities, and it can easily integrate a third-party system for intent recognition. DialogFlow was integrated, for instance, to recognize complex intents corresponding to the user's memories, preferences, norms, and beliefs in \cite{grassi2022}. 

In June 2020, the OpenAI company released an API\footnote{\url{https://openai.com/blog/openai-api/}} that provides access to GPT-3\footnote{\url{https://en.wikipedia.org/wiki/GPT-3}}. Unlike the aforementioned conversational agents, GPT-3 is a generative model based on human-human data that uses deep learning to produce human-like utterances fitting the context, allowing an “open" conversation instead of working exclusively in Q\&A scenarios. Currently, most commercial generative models are either completely inaccessible to the public, accessible only for online testing and training\footnote{\url{https://openai.com/blog/chatgpt/}}, or gated by an API. Among the top open source language models there are GPT-J\footnote{\url{https://huggingface.co/EleutherAI/gpt-j-6B}}, GPT-NeoX \cite{gpt-neox}, and BLOOM (BigScience Large Open-science Open-access Multilingual Language Model)\footnote{\url{https://huggingface.co/bigscience/bloom}}. 

Despite their promising capabilities, all generative models have the major drawback that they can learn undesirable features leading to toxic or biased language \cite{mcguffie2020}. This drawback makes them unsuitable in sensitive contexts, such as the development of socially assistive robots taking care of frailer people or in any context when cultural sensitivity may be required. Moreover, open-source language models need a very high-performance computer. Suffice it to say that the download size of BLOOM is about 700GB, and it needs more than 350GB RAM.
Finally, even if generative models can be used in many ways (Q\&A, text completion, grammar correction, translation, code completion, etc.), conversational agents based on these models are not designed to manage a complex, human-like dialogue with a person. Generative models do not address the problem of taking the initiative or engaging the person; they never express opinions, pose questions, or propose activities. Indeed, this is the objective of CAIR, which may use these models as third-party services to improve its versatility (a possibility we are currently exploring).

\subsubsection{Planning}
In addition to having conversations with the users, social agents should also be able to model and reason about more complex tasks or activities to be carried out in cooperation with other agents. Activity planning has been extensively analyzed in literature when dealing with a single agent embedded with the required onboard capabilities. Recently, cloud technologies have been used to improve planning capabilities: the computing resources of the cloud can be used to perform more complex computations and gather information that can be useful to all agents connected.

A cloud-based system architecture for robotic path planning is presented in \cite{lam2014}. The cloud server contains a path plan database, which stores the solution paths that can be shared among robots. The system also provides on-demand path planning software in the cloud, which computes the optimal paths for robots to reach the goal positions. The authors have experimentally verified the feasibility and effectiveness of solving the shortest path problem via parallel processing in the cloud. 
An example of task planning in a dynamic global environment framework is presented in \cite{joo2019}. The work proposes a cloud-based framework for real-time autonomous robot navigation with 3D visual semantic SLAM that exploits on-demand databases to store environment information. Another planning example, aimed at managing a fleet of autonomous mobile robots (AMR) using Rapyuta Cloud Robotics Platform, is provided in \cite{singhal2017}, whereas \cite{zagradjanin2019} discusses a multi-robot system based on cloud technologies, designed to execute tasks in a complex and crowded environment. The RoboEarth cloud engine \cite{Riazuelo2015RoboEarthSM} includes a database to store information that can be used in several different scenarios, including action recipes and skills, speeding up robot learning and adaptation in complex tasks. In all the examples above, the cloud approach shifts the computation load from the agents to the cloud and provides powerful processing capabilities to the multi-robot system.

In its current development stage, CAIR does not provide advanced capabilities for planning, as it basically searches for plans that match the user's goals. Its main feature is the availability of a formalism to describe both plans to operate in the real world (in a way that is abstracted from the actual agent's sensorimotor capabilities and requires to be grounded into such capabilities during execution), and plans to operate on the knowledge that the robot has about the person, its view of the world, and the state of the conversation. Integration with third-party planners is possible and currently being explored.

\subsubsection{Collective learning}
Collective learning refers to the sharing, storing, and accumulating of information over time, a capability that allows social agents to work together efficiently. Examples of information that agents can share for collective learning are control policies, sensor information on physical traits of an environment, trajectories, tracking data, and updated localization data. 

To enable robots to perform human-level tasks flexibly in varying conditions, \cite{kaan2018} argues that we need a mechanism that allows them to exchange knowledge. One approach to achieve this is to equip a cloud application with a range of encyclopedic knowledge in the form of an Ontology and execution logs of different robots performing the same tasks in different environments. In a similar spirit, \cite{chibani2013} and \cite{sandygulova2013} describe a collective learning environment for ubiquitous robots. Sensors embedded in these robots can provide vast amounts of information that can benefit further processing and collective learning. The RoboEarth three-layered architecture \cite{Riazuelo2015RoboEarthSM} emphasizes that each robot should allow other robots to learn through its knowledge and vice-versa. Thanks to a portfolio of web services, RoboEarth allows access to a database storing information that can be reused in several different scenarios, including images, point clouds, models, maps, and object locations. 
Among applications, \cite{bonaccorsi2016} investigates and assesses how a cloud robotic system can improve the provisioning of assistive services to promote active and healthy aging. In this scenario, the presence of a cloud robotic service is fundamental to design agents that can simultaneously monitor more elderly people at the same time, regardless of the time and the location of the seniors, through sensors located in the environment. 

Ongoing work is being done to integrate an approach for crowdsourcing new knowledge during user interactions into CAIR, as documented in \cite{grassi2022}. However, this aspect will not be discussed further in the following.

\subsubsection{Perception}
Perception assumes significant importance for human-robot interaction. It is reasonable to identify four main classes of signals captured by a social robot: visual-based, audio-based, tactile-based, and range sensors-based. Robots collect such data through cameras, microphones, tactile sensors, and proximity sensors such as laser range finders, ultrasounds, infrared, or even RGB-D cameras. Semantic understanding includes processing and merging of sensor data for tasks such as speech-to-text translation, sound localization, natural language understanding, activity, gesture, posture, emotion recognition, object localization and recognition, and many others. 

According to this rationale, \cite{soyata2012} and \cite{pawle2013} discuss the design and implementation of face recognition applications exploiting the benefits of cloud computing. The popularity of handy smart devices allows healthcare providers to monitor patients’ health without visiting them. As proof, the work described in \cite{hossain2015} proposes a cloud-assisted speech and face recognition framework for elderly health monitoring, where handheld devices or video cameras collect speech, along with face images, and deliver them to the cloud server for possible analysis and classification. A cloud-based framework for speech-enabling healthcare is proposed in \cite{muhammad2015}. A person seeking some medical assistance can send their request by speech commands to a cloud server where such requests are managed and processed. Any doctor with proper authentication can receive the request and assist the person.

Advanced capabilities for perceiving the user's emotions \cite{Demutti20221435} are currently being integrated into CAIR and therefore are not discussed in this work. 

\section{System Architecture}
\label{sec:sys-arch}
This section describes the system architecture, starting with a general overview, then detailing the implementation of the cloud server and explaining the basic operations a client should perform.

\subsection{General Principles}

\begin{figure}
    \centering
    \includegraphics[width=\linewidth]{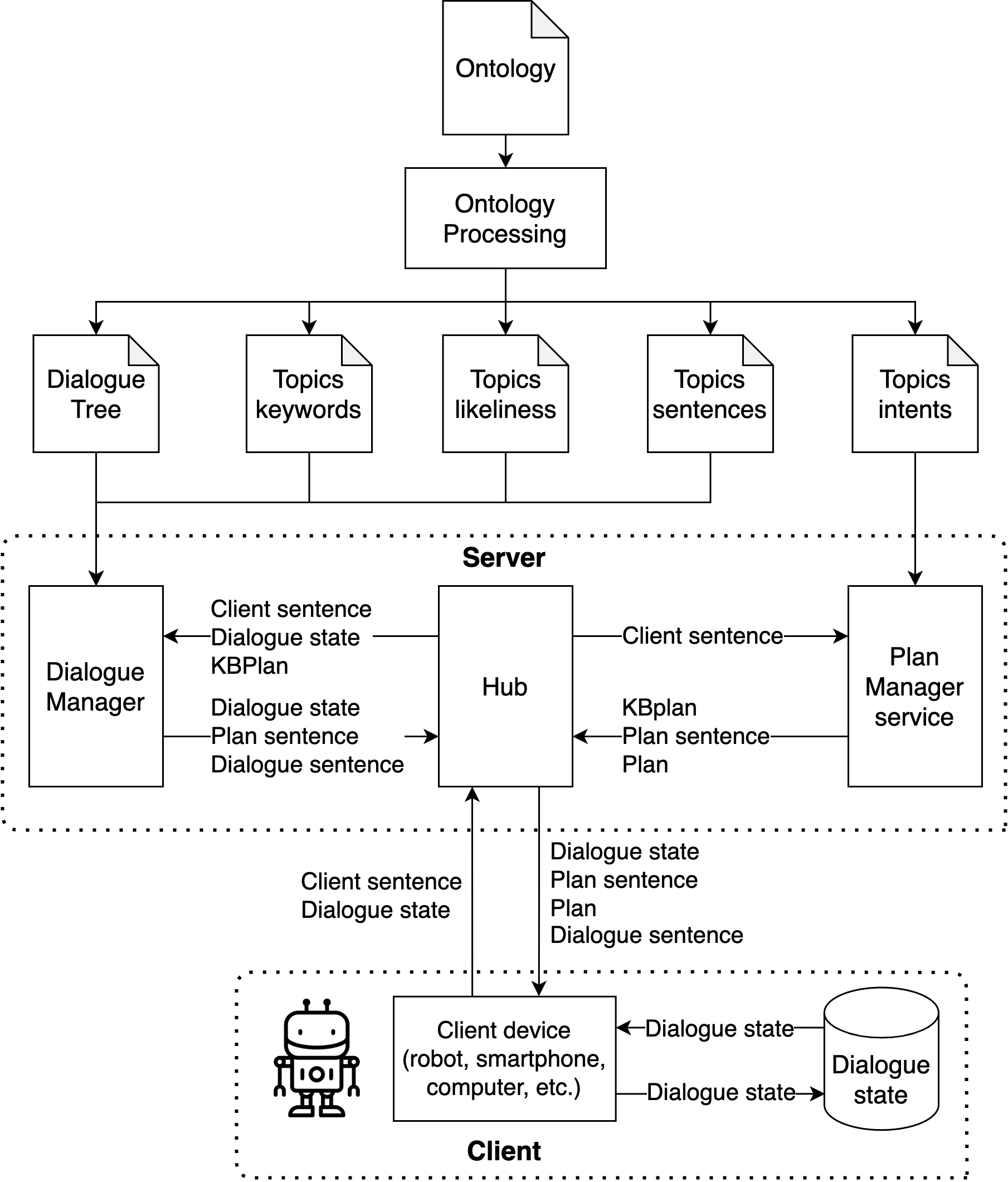}
    \caption{CAIR system architecture.}
    \label{fig:architecture}
\end{figure}

Figure \ref{fig:architecture} depicts the architecture of the system. The cloud server is composed of three web services developed in Python: (1) the Hub (Section \ref{sec:hub}) that manages the incoming requests from the client, (2) the Plan Manager (Section \ref{sec:plan}) that recognizes the intention of the user to make the agent execute a task, and (3) the Dialogue Manager (Section \ref{sec:dialogue}) that manages the dialogue and recognizes the user's intention of talking about a specific topic. To provide appropriate answers and plans, the cloud server exploits an Ontology \cite{guarino1998} implemented in OWL 2 \cite{motik2008} containing all the topics, sentences, and plans used during the interaction with the user.

The Flask-RESTful\footnote{\url{https://flask-restful.readthedocs.io/en/latest/}} framework is used to develop the web services. The client can perform requests to the cloud server using REST APIs (Representational State Transfer), a set of rules that should be followed when creating the API \cite{masse2011}. One of these rules states that the client should be able to get a piece of data (called a \textit{resource}) when linked to a specific URI (Uniform Resource Identifier). Any web service that obeys the REST constraints is informally described as RESTful. Due to the separation between client and server, this protocol allows developers to use different syntaxes on different platforms.

As shown in Figure \ref{fig:architecture}, a client running on a robot or a device simply has to acquire the sentence pronounced by the user, send it to the Hub service along with the \textit{dialogue state}, parse the response, store the updated \textit{dialogue state}, execute the received \textit{plan}, and/or reply with the \textit{dialogue sentence} returned by the Hub. How these operations are performed depends on the capabilities of the client. The acquisition of the user's sentence can be performed both through a keyboard or a microphone. The \textit{plan} will be performed only if the device running the client has the appropriate physical capabilities; otherwise, it will be ignored. Eventually, the \textit{dialogue sentence} will be displayed through a screen, or it will be communicated to the user through a speaker (assuming that the device is also equipped with a voice synthesizer). 

\begin{remark}
The purpose of the \textit{dialogue state} is to exchange updated information about the client between the client and the server, while preventing the storage of sensitive information on the cloud server.
\end{remark}

\subsection{Hub}
\label{sec:hub}
The Hub service is the facade of the system, and it is designed to receive all the requests from clients.

\begin{figure}
    \centering
    \includegraphics[width=\linewidth]{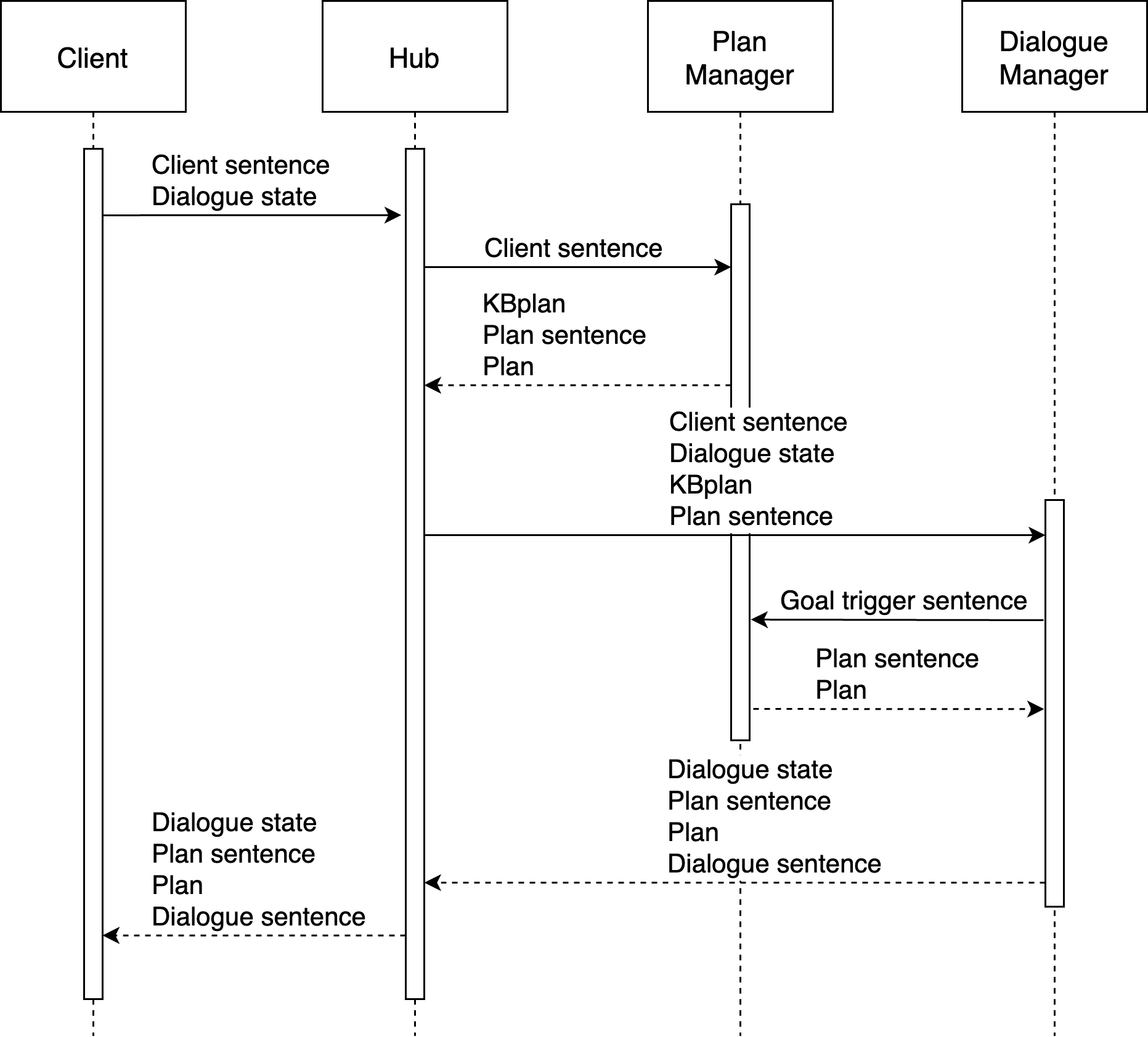}
    \caption{Sequence diagram showing the interaction between the client and the web services on the CAIR server.}
    \label{fig:sequence_diagram}
\end{figure}

The diagram in Figure \ref{fig:sequence_diagram} shows the sequence of the interactions between the client and the web services on the cloud. The \textit{client sentence} pronounced by the user will be processed in the pipeline by the Plan Manager\footnote{This service could be directly called by the client if the user is not interested in managing the dialogue.} (which aims to recognize the user's intents) and the Dialogue Manager (which aims to move the conversation forward). The Plan Manager will return a \textit{plan} or a \textit{KBplan} (and an appropriate introductory sentence to the plan, \textit{plan sentence}) if and only if the intent is recognized, whereas the Dialogue Manager will always produce a \textit{dialogue sentence} that the agent may say to move the conversation forward. The \textit{dialogue state} is sent by the client to the server together with the original \textit{client sentence}, updated, and sent back to the client. 

\subsubsection{Request to Plan Manager}
The first operation that the Hub performs is a request to the Plan Manager service to check if the \textit{client sentence} pronounced by the user matches any of a pre-defined set of intents associated with specific plans to be executed. These plans could either represent a sequence of actions to be executed by the client or affect the knowledge base and/or the flow of the dialogue (in this case, referred to as a \textit{KBplan}). If the sentence matches an intent, the service will return the  \textit{plan} or the \textit{KBplan}, along with an introductory \textit{plan sentence} to be used by the client before performing the first action in the plan (the difference between \textit{plan} and \textit{KBplan} is explained more in detail in Section \ref{sec:plan}). After that, the \textit{KBplan} (if any) 
will be collected together with the original \textit{client sentence} and the \textit{dialogue state} to perform a request to the Dialogue Manager service. 

\subsubsection{Request to Dialogue  Manager}
The second operation that the Hub performs is a request to the Dialogue Manager, which will reply with a \textit{dialogue sentence} to move the conversation forward. Figure \ref{fig:sequence_diagram} shows that, in this process, the Dialogue Manager has an additional way to trigger a plan: if the original \textit{client sentence} contains a positive reply to an activity proposed by the system in the previous turn, the Dialogue Manager will query the Plan Manager, which produces a corresponding \textit{plan} and \textit{plan sentence}. 

\subsubsection{Reply to Client}
Eventually, as the third and last operation, the Hub collects all the information provided by the Plan Manager and the Dialogue Manager. Such a set of data is finally returned as a response to the client, which will implement it through its sensorimotor (\textit{plan}) and verbal capabilities (\textit{dialogue sentence} and/or \textit{plan sentence}).

\subsection{Plan Manager}
\label{sec:plan}
\begin{figure}[!ht]
    \centering
    \includegraphics[width=0.97\linewidth]{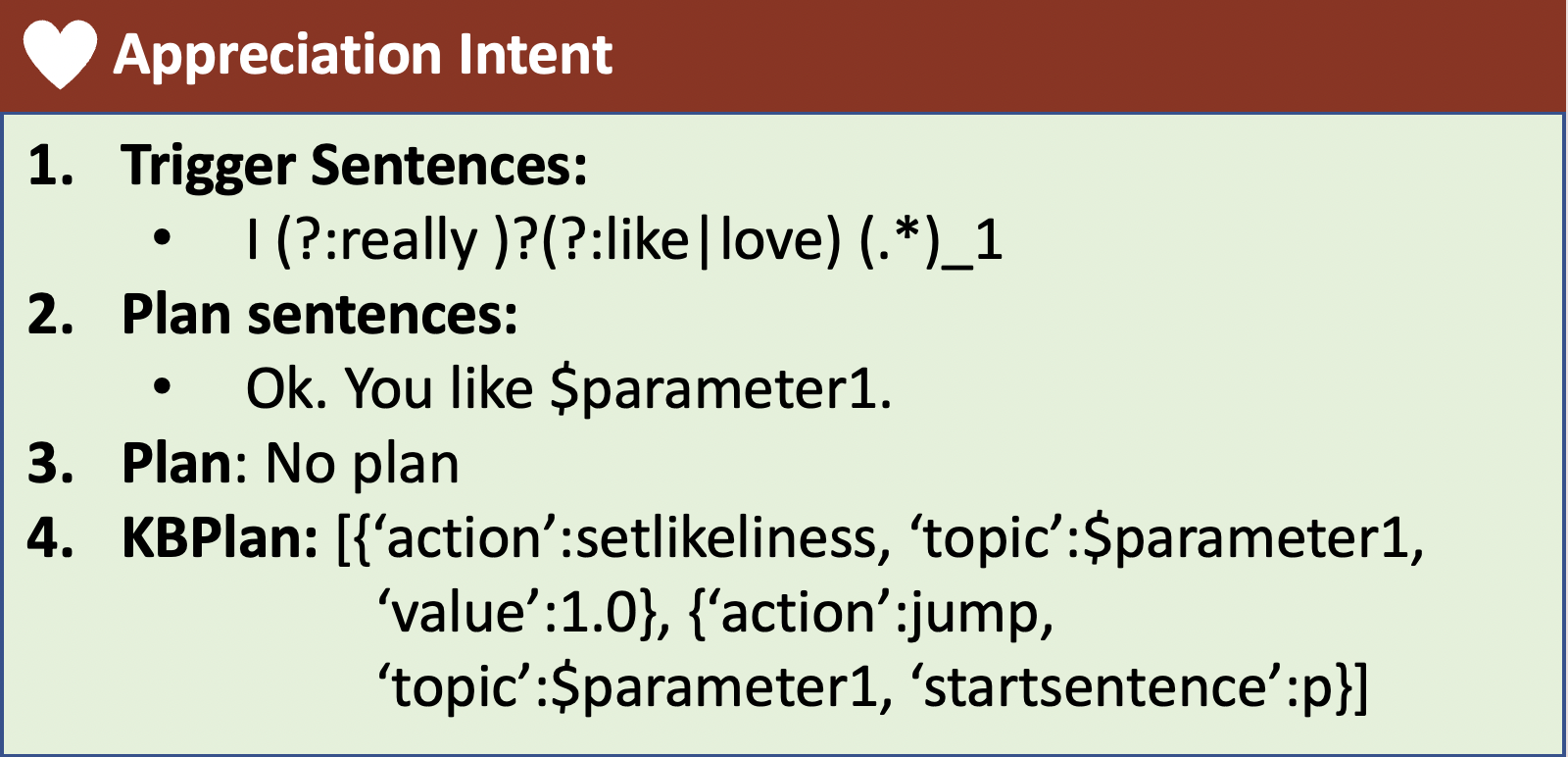}
    \includegraphics[width=0.97\linewidth]{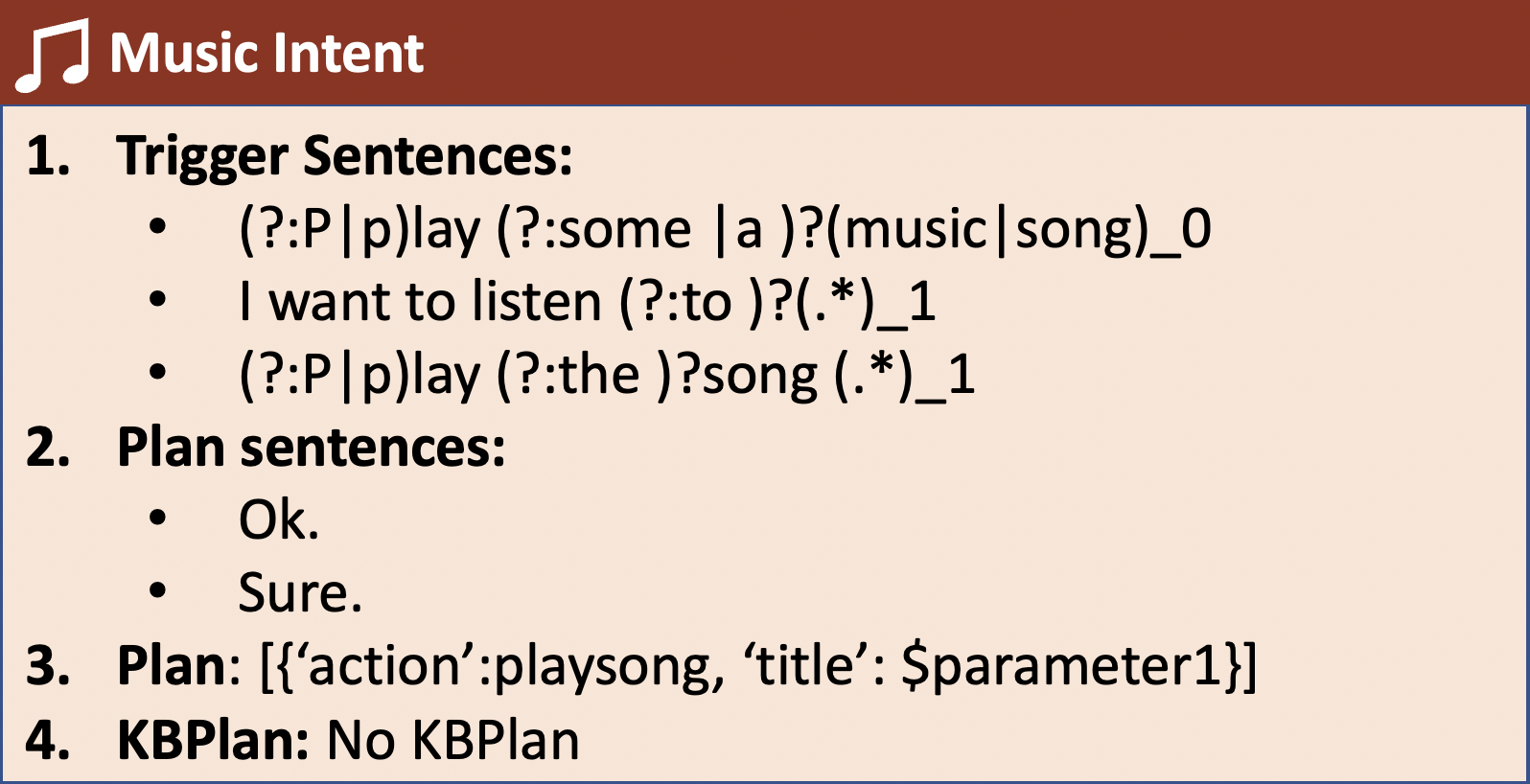}
    \includegraphics[width=0.97\linewidth]{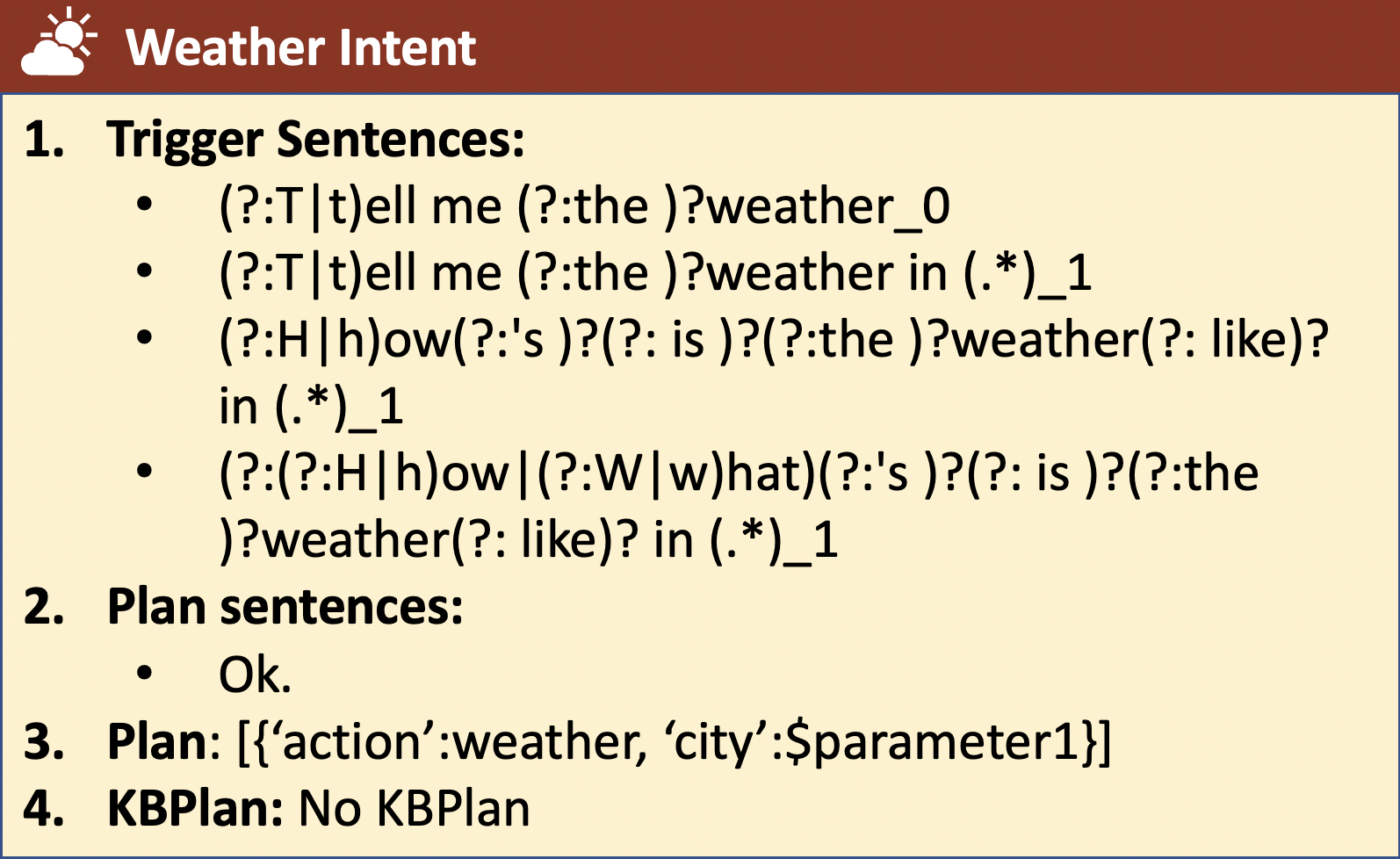}
    \includegraphics[width=0.97\linewidth]{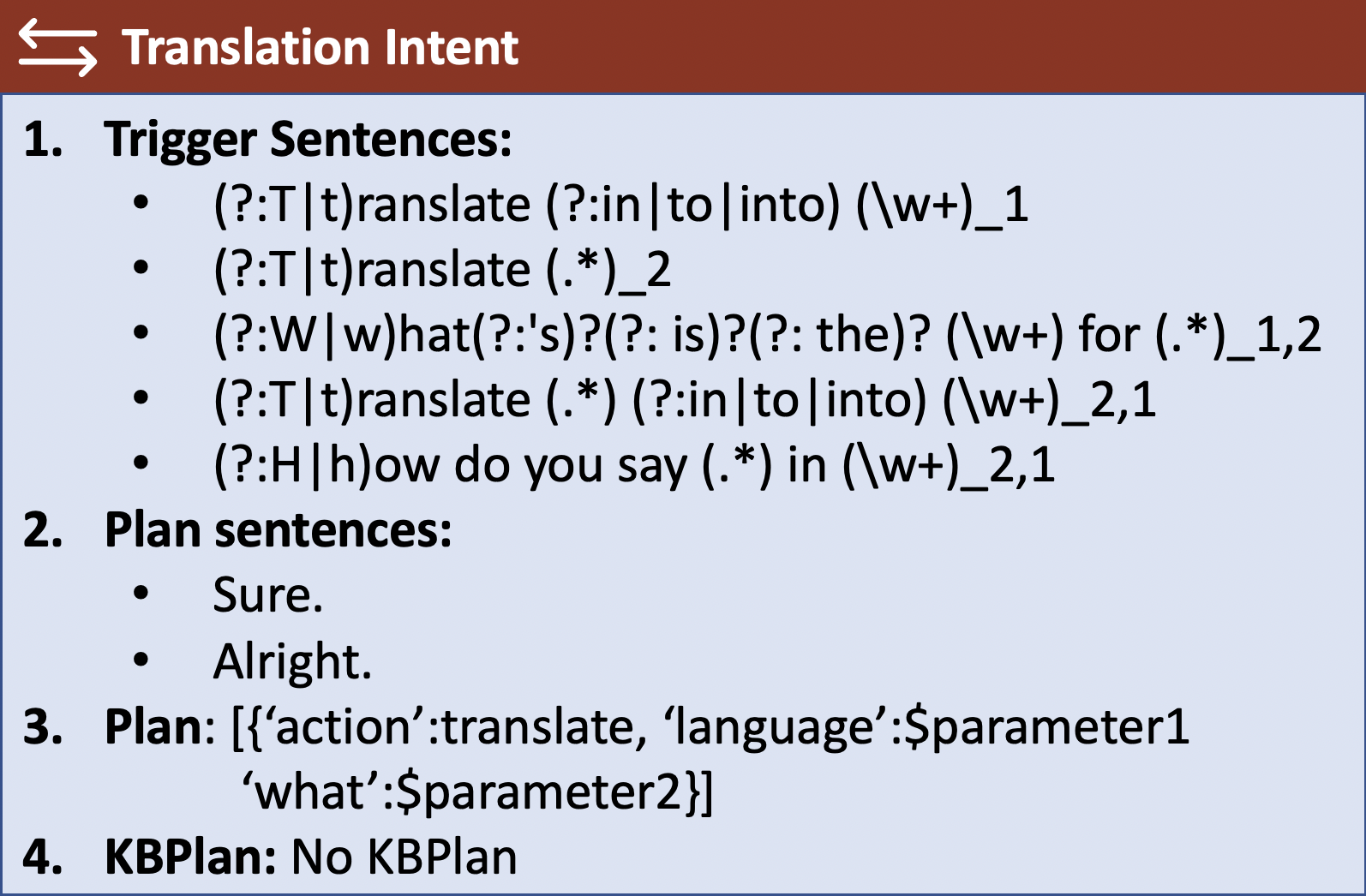}
    \caption{Examples of intents recognized by the Plan Manager service.}
    \label{fig:intents}
\end{figure}

The Plan Manager service receives as input the \textit{client sentence} pronounced by the user. Its purpose is to find a match between such a sentence and one of the trigger sentences associated with a specific intent. An intent is defined by (i) a set of trigger sentences, (ii) one or more \textit{plan sentences}  (if any), (iii) a \textit{KBplan} (if any), (iv) and a \textit{plan} (if any). Figure \ref{fig:intents} shows some examples of intents that can be recognized by the system. 

In the current implementation, sentence matching is performed based on pattern-based syntax matching and allows for the extraction of parameters from the matched sentence that can be used to dynamically compose the \textit{plan sentences}, the \textit{plan}, and the \textit{KBplan}. More complex models merging syntax matching and data-driven example-based approaches \cite{mrkvsic2017} can be easily integrated without any impact on the general structure.

\subsubsection{KBplan}
A \textit{KBplan}, where KB stands for Knowledge Base, is a sequence of actions meant to affect the knowledge base and/or the flow of the dialogue. For instance, if the user says \textit{``I love music"}, this sentence will match the trigger sentences of the Appreciation Intent (Figure \ref{fig:intents}) meant to recognize the user's appreciation for something and extract the loved thing as a parameter. The \textit{KBplan} of this intent is composed of two actions: the first one is meant to modify the probability that the user wants to talk about the extracted parameter (what we will refer to as ``likeliness" associated with a topic in the Ontology in the following Section), while the second one brings the information that the system should jump to that conversation topic (if present in the Ontology).

\subsubsection{Plan}
A \textit{plan} is a sequence of actions that should be executed on the client (given that it has appropriate capabilities to execute it) as it does not affect the knowledge base or the flow of the dialogue. For instance, if the user says \textit{``Play the song Hey Brother"}, this sentence will match one of the trigger sentences of the Music Intent (Figure \ref{fig:intents}) that recognizes the user's intention to listen to some music. The \textit{plan} associated with this intent is composed of a single action carrying the information that the client should play the song having the title contained in the parameter field (see Section \ref{sec:client}). Other plans we considered may require the robot to move from one place to another (depending on its sensorimotor capabilities), greet the user, learning the name and position of relevant areas in the house, and others. 

Figure \ref{fig:architecture} shows that, if the Plan Manager finds a match with an intent, a response containing the \textit{KBplan}, the \textit{plan sentence}, and the \textit{plan} is returned to the Hub service: the \textit{KBplan} can be empty if the matched intent is associated only with a \textit{plan}, and the vice-versa can also be true.

\begin{remark}
Triggered intents might be interpreted as goals to be achieved by planning a sequence of actions in run-time, instead of having pre-cooked sequences of actions as in the current version of the system. Thanks to the versatility of RESTful APIs, we may easily add a planner based on the most popular formalisms such as PDDL \cite{fox2003}, Hierarchical Task Networks \cite{erol1994, hayes2016}, Answer Set Programming \cite{lifschitz2002, erdem2012}, Partially Observable Markov Decision Process \cite{kaelbling1998, ross2008} without altering the general structure. In this way, a sequence of actions might be produced considering constraints related to the actual capabilities of the physical device that needs to implement them, including those related to interaction with humans \cite{mumm2011, chen2017}. 
\end{remark}

\subsection{Dialogue Manager}
\label{sec:dialogue}
The Dialogue Manager service is in charge of managing the dialogue.

\subsubsection{Knowledge Representation}
\label{Knowledge Representation}
Following previous work \cite{bruno2019, recchiuto2020, papadopoulos2020}, the system has been designed with the capability of conversing with the users taking into account their cultural background, by relying on a framework for cultural knowledge representation based on an OWL 2 Ontology. According to the Description Logics formalism, concepts (i.e., conversation topics that the system is capable of talking about) and their mutual relations are stored in the terminological box (TBox) of the Ontology. Instead, instances of concepts and their associated data (e.g., chunks of sentences automatically composed to enable the system to talk about the corresponding topics) are stored in the assertional box (ABox). 

\begin{figure}[tp]
    \centering
    \includegraphics[width=\linewidth]{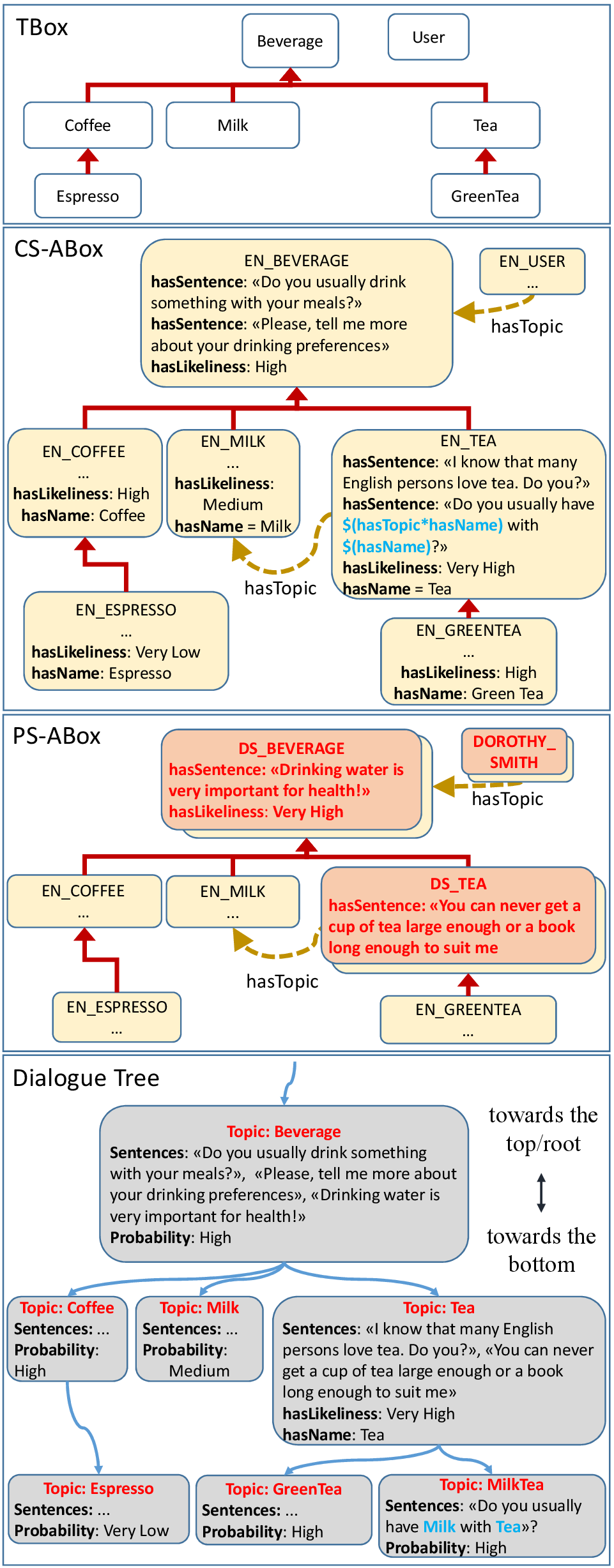}
    \caption{The three layers of the Ontology: TBox, CS-ABox (for the English culture), PS-ABox (for Mrs. Dorothy Smith), and the Dialogue Tree generated from the Ontology structure.}
    \label{fig:three-layers}
\end{figure}

To deal with representations of the world that may vary across different cultures \cite{carrithers2010}, the Ontology is organized into three layers (as explained more in detail in \cite{recchiuto2020, bruno2019}). The TBox (Figure \ref{fig:three-layers}) encodes concepts at a generic, culture-agnostic level and includes concepts that exist across all cultures considered, whichever the cultural identity of the user is, to avoid stereotypes. Consider beverages: the system may initially guess the user's preferred beverages depending on the country they live in. Nevertheless, it should be open to considering choices that may be less likely for a given culture, as the user explicitly declares their attitude towards them. For instance, the system may initially infer that an English person may be more interested to talk about \texttt{Tea} rather than \texttt{Coffee}, and the opposite may be initially inferred for an Italian user. However, during the conversation, initial assumptions may be revised. This mechanism leads to a stereotype-free, fully personalized representation of the user's attitude towards all concepts in the TBox to be used for conversation.

To implement this mechanism, the Culture-Specific ABox layer (CS-ABox in Figure \ref{fig:three-layers}) contains instances of concepts encoding culturally appropriate chunks of sentences, suggested by experts, to be automatically composed (Data Property \texttt{hasSentence}) and the probability that the user would have a positive attitude towards that concept, given that they belong to that cultural group. This idea has already been introduced in the previous section, where we used the term “likeliness" referring to the probability of having a positive attitude towards a concept in the Ontology (Data Property \texttt{hasLikeliness} with a value ranging from 0 to 1). 

Eventually, the Person-Specific ABox (PS-ABox in Figure \ref{fig:three-layers}) comprises instances encoding the actual user’s attitude towards a concept updated during the interaction (the system may discover that Mrs. Smith is more familiar with having tea than the average English person, instance \texttt{DS\_TEA} with the \texttt{hasLikeliness} property having a high value). During the first encounter between the robot and a user, many instances in the Ontology will not contain Person-Specific knowledge: the robot will acquire awareness about the user's attitude at run-time, either from its perceptual system or through verbal interaction, e.g., asking questions.

\subsubsection{Dialogue Tree}
A Dialogue Tree (DT) (Figure \ref{fig:three-layers}), used by the conversation system to chit-chat with the user, is built starting from the Ontology structure: each concept of the TBox and the corresponding instances of the ABox are mapped into a conversation topic, i.e., a node of the tree. The Object Property \texttt{hasTopic} and the hierarchical relationships among concepts and instances are analyzed to define the branches of the DT. In the example of Figure \ref{fig:three-layers}, the instance of \texttt{Tea} for the English culture is connected in the DT to its child node \texttt{GreenTea} (which is a subclass of \texttt{Tea} in the Ontology) and its sibling \texttt{MilkTea} (since \texttt{EN\_MILK} is a filler of \texttt{EN\_TEA} for the Object Property \texttt{hasTopic}). 

Each conversation topic is associated with the following:

\begin{itemize}
\item Chunks of culturally appropriate sentences associated with the topic that are automatically composed and used during the conversation. Such sentences can be of different types (i.e., yes/no questions, positive or negative assertions, open questions, activity proposals, etc.) and may contain variables that are instantiated when the Dialogue Tree is created;
\item A specific sequence in which sentences of different types are used by the system to produce an engaging conversation flow. For example, the system could begin by asking a yes/no question to measure the user's interest in the topic. If the user is interested, the system could provide a positive assertion about the topic; otherwise, it may provide a negative assertion. Following that, an open-ended question could be used to obtain more information about the user's thoughts or previous experiences related to the topic. Finally, the system may propose an activity.  
\end{itemize} 

To clarify the mechanism to compose sentences, suppose,a hypothetical sentence “Do you like \$hasName?": when encoded in the concept \texttt{Coffee} might be used to produce both “Do you like Coffee?" and “Do you like Espresso?" (being \texttt{Espresso} a subclass of \texttt{Coffee}). In the basic version used for testing, the taxonomy of the Ontology allowed us to easily produce a DT with about $3,000$ topics of conversation and more than $20,000$ sentences, with random variations made in run-time. The reason for using a “safe" Ontology-based mechanism, in which chunks of sentences have been written by experts rather than a generative model, has already been explained in Section \ref{sec:SOTA}.


To implement the knowledge-based conversation mechanism, the Dialogue Manager needs the Dialogue Tree with rules to compose sentences, the \textit{client sentence}, and the \textit{dialogue state}. 
As described more in detail in Section \ref{sec:client}, the \textit{dialogue state} contains the current conversation topic, the type of the previous 
and the following sentence in the ordering, the likeliness of all topics mentioned in the conversation up to that moment (about which the user may have expressed a preference), and the sentences that have already been used (to reduce unnecessary repetitions).

Given these premises, and based on the Dialogue Tree, the key ideas for knowledge-driven conversation are summarized in Algorithm \ref{algorithm1} (the whole process \cite{recchiuto2020} is more complex). 

\subsubsection{Execution of KBplans}
Firstly, recall that the Hub queries the Plan Manager before the Dialogue Manager. Hence, when the Dialogue Manager is queried, it receives not only the \textit{client sentence} and the \textit{dialogue state} but possibly the \textit{KBplan} returned by the Plan Manager (if any). Each time the Dialogue Manager receives them from the Hub, it first checks if the \textit{KBplan}  contains at least an action to operate on the Ontology (lines $2$ to $20$ of Algorithm \ref{algorithm1}). For instance, the \textit{KBplan} might contain the ``setlikeliness" (line $4$) and the ``jump" action (line $9$) reported in the Appreciation Intent in Figure \ref{fig:intents}. 

Line $16$ summarizes additional actions that might be part of a \textit{KBplan}. 

To show how a \textit{KBplan} is executed, consider the ``setlikeliness" action, corresponding to an explicit attempt of the user to declare their attitude towards a topic: when the user says ``I love music," the action tells the Dialogue Manager to update the likeliness of the concept ``music" (if it exists in the Ontology). Note that the likeliness is modified only for that specific client/user and that the cloud server does not store any information about clients/users: the likeliness of the topic is updated only in the \textit{dialogue state} sent back and forth between the client and the cloud server, 
(line $7$). 
The ``jump" action, the second step of the Appreciation Intent, 
requires the Dialogue Manager to change the conversation topic to the one specified in the ``topic" field of the action (if it exists in the Ontology) and to choose a sentence associated with the new topic. 

\scalebox{0.80}{
    \begin{minipage}{1.0\linewidth}
\begin{algorithm}[H]
	\begin{algorithmic}[1]
	\Require{client\_sentence, dialogue\_state, KBplan}
	    \State{topic $\gets$ dialogue\_state.topic}
	    \If{HasElements(KBplan)}
	        \ForAll{action in KBplan.actions}
	            \If{action = ``setlikeliness"}
	                \State{new\_topic $\gets$ action.topic }
    		        \If{HasTopic(Ontology, new\_topic)}
    		            \State{dialogue\_state.topic.likeliness $\gets$ UpdateLikeliness(new\_topic, action.value)}
    		        \EndIf
	            \ElsIf{action = ``jump"}
	                \State{new\_topic $\gets$ action.topic}
	                \If{HasTopic(Ontology, new\_topic)}
    	                \State{dialogue\_sentence $\gets$ ChooseSentence(new\_topic, action.startsentence)}
                        \State{dialogue\_state $\gets$ UpdateState(new\_topic, ...)}
    		            \State \Return dialogue\_sentence, dialogue\_state
		            \EndIf
		        \ElsIf{action = $<$some action$>$}
		            \State{$<$do something$>$}
		        \EndIf
		       \EndFor
	    \EndIf
	    \State {keywords $\gets$ RetrieveKeywords(client\_sentence)}
	    \State {matched\_topics $\gets$ SearchTopic(Ontology, keywords)}
	    \If{HasElements(matched\_topics)}
            \State{new\_topic $\gets$ OneOf(matched\_topics)}
            \State{dialogue\_sentence $\gets$ ChooseSentence(new\_topic, new\_topic.first\_sentence\_type)}
            \State{dialogue\_state $\gets$ UpdateState(new\_topic, ...)}
		    \State \Return dialogue\_sentence, dialogue\_state
		\Else
		    \If{Completed(topic)}
                \State{new\_topic $\gets$ Next(DT, topic)}
                \State{dialogue\_sentence $\gets$ ChooseSentence(new\_topic, new\_topic.first\_sentence\_type)}
                \State{dialogue\_state $\gets$ UpdateState(new\_topic, ...)}
                \State \Return dialogue\_sentence, dialogue\_state 
		    \Else
		        \State{dialogue\_sentence $\gets$ ChooseSentence(topic, topic.next\_sentence\_type)}
                \State{dialogue\_state $\gets$ UpdateState(topic, ...)}
		        \State \Return dialogue\_sentence, dialogue\_state
		    \EndIf
		\EndIf
	\caption{Dialogue Manager Algorithm}
    \label{algorithm1}
	\end{algorithmic}
\end{algorithm}
\end{minipage}%
}

Since there are different sentences of different types for each topic with a defined ordering, the type of the sentence to be selected is specified in the ``startsentence" field of the action (line $12$). As an example, in the Appreciation Intent shown in Figure \ref{fig:intents}, the ``startsentence" parameter of the ``jump" action is ``p", which means that the system will choose one among the \textit{positive} (affirmative) sentences associated with the topic (e.g., “Music is good for your health!”). Remember that the topic has been extracted by the Plan Manager starting from the \textit{client sentence}, and the sentence is automatically composed starting from chunks of sentences validated by experts for their appropriateness.

After updating the \textit{dialogue state} (line $13$) with the new topic and all the related information (i.e., likeliness, sentences that have already been used by the system, and the type of the following sentences to be used for that topic), the algorithm returns the chosen \textit{dialogue sentence} and the updated \textit{dialogue state} (line $14$). The \textit{dialogue state} will be stored on the client side and sent again to the server during the next interaction. 

\subsubsection{Managing the conversation flow}
In case the \textit{KBplan} does not contain any element (lines $21$ to $39$ in \ref{algorithm1}), the Dialogue Manager algorithm checks if the \textit{client sentence} contains keywords encoded in the Ontology in a corresponding Data Property. To match a topic, at least two keywords associated with that topic should be detected in the \textit{client sentence} pronounced by the user (one of which can be a wildcard for more versatility in keyword matching). The use of multiple keywords allows the system to differentiate between semantically close topics (i.e., \texttt{Green Tea} rather than the more general concept of \texttt{Tea}). 

If the \textit{client sentence} contains the two keywords corresponding to at least a topic in the Ontology (line $23$), the algorithm jumps to one of the matching topics to proceed with the dialogue (matching topics are randomly chosen based on their likeliness, i.e., how much the topic matches the user's cultural and personal preferences).
Specifically, every time the conversation topic changes, the system starts talking about the new topic by picking a sentence of the first type according to the ordering (line $25$), e.g., a yes/no question. The \textit{dialogue state} is updated and returned along with the chosen \textit{dialogue sentence} (lines $26$ and $27$). 

Otherwise, if no keywords are detected, the current topic is further explored as follows:
\begin{itemize}
    \item If the topic has been completely explored, i.e., all sentence types associated with that topic have been considered in the ordering 
    (line $29$), the next topic is chosen either by exploring the DT along its branches (i.e., favoring semantically close topics, if any) or choosing a completely different topic close to the root of the DT 
    (line $31$). The \textit{dialogue state} and a \textit{dialogue sentence} are returned (lines $32$ and $33$).
    \item If the topic has not been completely explored, i.e., there are still sentence types associated with that topic to be considered in the ordering 
    (line $34$), the system stays on that topic. While in the same topic, a random sentence of the required type is chosen following the ordering at each iteration. 
    The \textit{dialogue state} and a \textit{dialogue sentence} are returned (lines $36$ and $37$).
\end{itemize}

As shown in Figure \ref{fig:sequence_diagram}, the Dialogue Manager can also directly perform a request to the Plan Manager service. For the sake of simplicity, this eventuality has not been reported in Algorithm \ref{algorithm1}. However, during the conversation, the system can also propose activities to the user, e.g., ``Do you want me to play some music?", which will ultimately trigger a plan in case the user answers affirmatively (proposals for activities are among the sentence types that can be chosen while exploring a topic). 

\begin{remark}
Similarly to what we did for intent matching, a more sophisticated mechanism for sentence matching might be adopted. This approach would merge syntax matching and data-driven example-based methods \cite{mrkvsic2017}, rather than relying solely on keyword matching.
\end{remark}

\subsection{Client}
\label{sec:client}
The client exchanges with the cloud server four types of information: \textit{dialogue state}, \textit{client sentence}, \textit{dialogue sentence}, and \textit{plan}.

\subsubsection{From the client to the cloud server}
\begin{figure*}[hbt!]
    \centering
    \includegraphics[width=\textwidth]{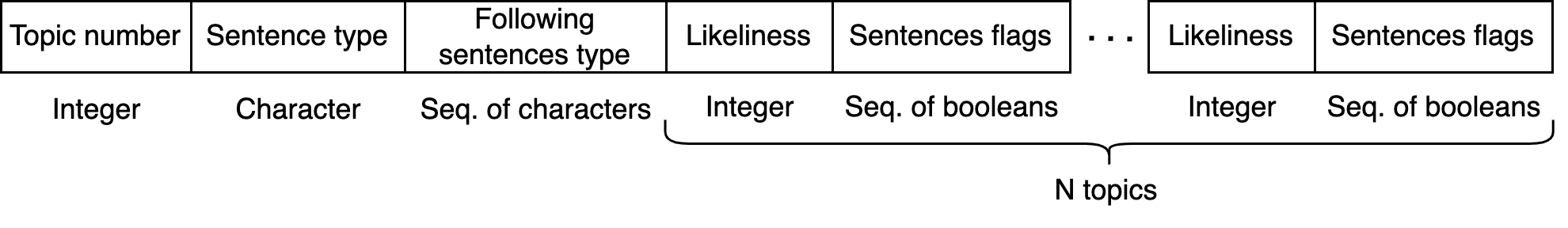}
    \caption{Information contained in the \textit{dialogue state}.}
    \label{fig:dialogue_state}
\end{figure*}

Every time the client interacts with the system, it should provide its \textit{dialogue state}, which will be updated by the server and transmitted back, together with other data. The \textit{dialogue state} is defined as the sequence of elements represented in Figure \ref{fig:dialogue_state}, 
which is encoded and compressed every time it is exchanged between the server and the client. 
Please notice that if a client has never interacted with the cloud server, the first thing that it should do is to perform a request to the cloud, in particular to the Hub service, to obtain the initial \textit{dialogue state} that will be stored locally.

As shown in the Figure, the \textit{dialogue state} contains information about:
\begin{itemize}
    \item The current conversation topic;
    \item The type of the last sentence chosen by the Dialogue Manager
    , which will be used to move the conversation forward;
    \item An ordered list containing the types of the next sentences that the server will use while in the current conversation topic;
    \item The set of the “likeliness" values that have been modified during previous conversations with a given user, encoding the user's preferences about covered topics (as explained in Section \ref{Knowledge Representation}, Person-Specific likeliness will override Culture-Specific likeliness depending on the user's cultural background);
    \item The sentences that have already been used by the Dialogue Manager, to limit repetitions of the same sentences.
\end{itemize}
Please note that, during the conversation, the size of the \textit{dialogue state} increases as the number N of covered conversation topics grows. This is because, as shown in Figure \ref{fig:dialogue_state}, the \textit{dialogue state} stores the likeliness and the already chosen sentences for each covered topic.

Finally, along with the initial \textit{dialogue state}, the client acquires the \textit{client sentence} pronounced by the user (e.g., through a text- or speech-based interface, depending on its physical capabilities) and sends them to the Hub. 

\begin{remark}
Thanks to this mechanism, we store all relevant information on the client, therefore creating a cloud server that is “stateless" in all possible senses. 
In addition to being beneficial in terms of privacy, since no personal data is stored on the server, this makes the whole RESTful architecture very efficient to debug. All services, possibly residing on different machines for load-balancing purposes, can be called at any time without any need for synchronization.
\end{remark}

\subsubsection{From the cloud server to the client}
As soon as the client receives a reply, it shall manage the response appropriately. This includes storing the updated \textit{dialogue state}, communicating the \textit{plan sentence} (if any) to the user, 
performing the actions contained in the \textit{plan} field of the response, and eventually continuing the dialogue by communicating the \textit{plan sentence}. If the client is not able to execute certain actions, it will ignore them and consider only the dialogue reply (e.g., the Pepper robot could perform the action “Go to the kitchen", while the Pillo robot could not; yet, Pillo can dispense pills). 

It is worth restating that the \textit{dialogue state} does not contain any personal information about the user (e.g., name, gender, phone number, etc.). Whenever needed, this information is stored locally on the client device and substituted to the placeholders in the \textit{dialogue sentence}. For instance, the placeholder \texttt{\$name} in the sentence “Hello \texttt{\$name}, how are you?" should be substituted on the fly with the locally-stored name of the user, before saying/visualizing the sentence.

An example of a simple client for PC, a documented example of a full client for the SoftBank Robotics robots Pepper and NAO, and a ROS2 wrapper for the client are available on request. 
The code is provided along with a guide with a detailed explanation of how it works and of all the plans that the server can return to the client based on the intent that has been matched.

A video showing some extracts of the interaction is available on YouTube\footnote{\url{https://youtu.be/hgsFGDvvIww}}. Another video, related to a trial performed at the “Parini-Merello" middle school and involving a total of 300 students, demonstrates the possibility of using the cloud system for autonomous multiparty interaction\footnote{\url{https://youtu.be/27YvsH7IoSA}}.

\section{Experiments}
\label{sec:experiments}
\begin{figure}[t]
    \centering
    \includegraphics[width=\linewidth]{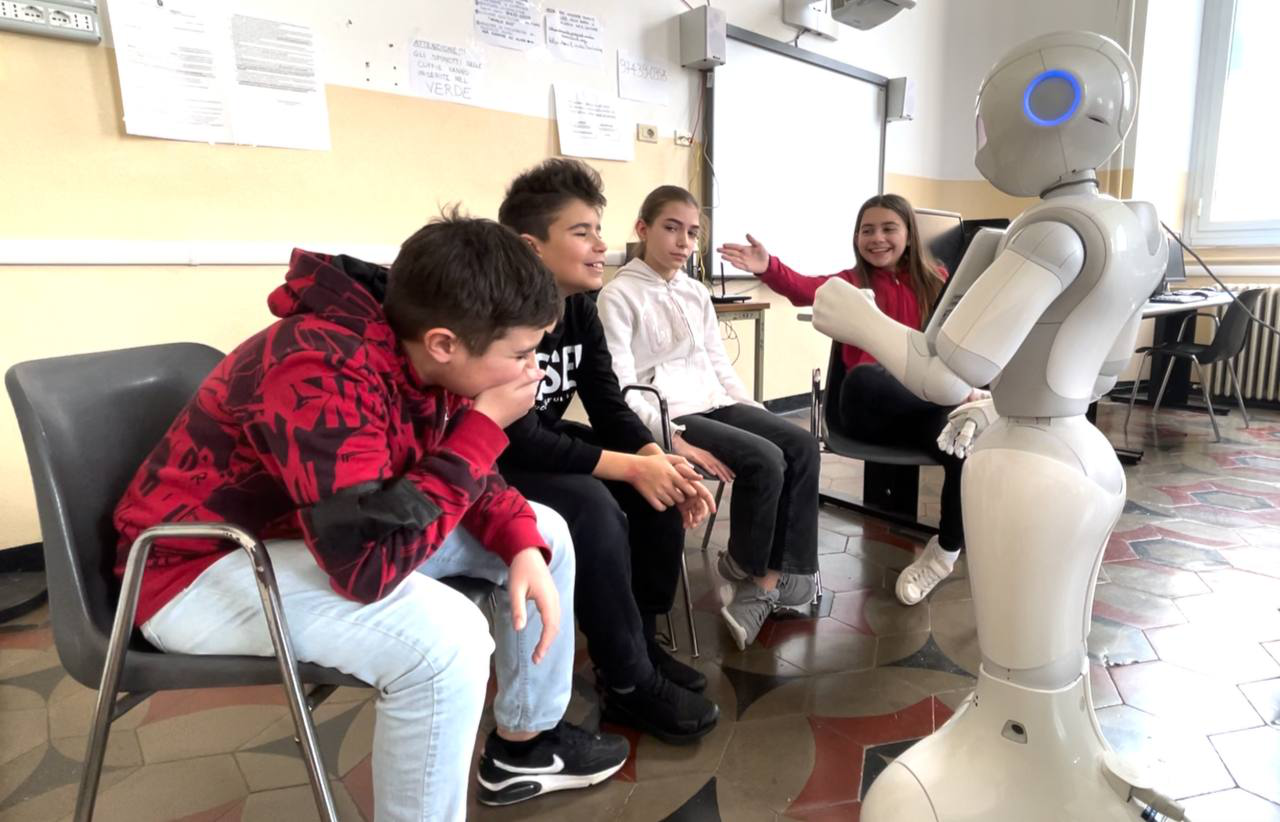}
    \caption{Middle-school students interacting with Pepper connected to the CAIR cloud (parents' consent to publication given).}
    \label{fig:lab_interaction}
\end{figure}

The system described in this paper is currently being used for research in Human-Robot Interaction (Figure \ref{fig:lab_interaction}).

As mentioned in the introduction, experiments to evaluate the system’s capabilities in terms of the quality of the interaction, usability, as well as the impact on life quality were performed with care home residents in the UK and Japan \cite{papadopoulos2020, Papadopoulos2021}, and additional trials have been performed with 300 high-school students (results under evaluation, Figure \ref{fig:lab_interaction}) and patients with severe spinal cord injuries (ongoing). The goal of the experiments described in this article is not user evaluation but to analyze the system performance in terms of response speed when multiple clients connect to the cloud in different configurations, paving the way to the deployment of culture-aware cloud services for robot scientists. 

Some initial considerations are necessary. From the experiments conducted in \cite{peng2020}, it emerges that users have the highest level of satisfaction with a maximum delay of two seconds during the conversation. As a confirmation of this, the empirical study in \cite{shiwa2008} about the response time of a communication robot revealed that users' evaluation of the robot's reply worsened after a response delay of two seconds. Also, we assume that, when interacting with our system, a user might perform, on average, six requests per minute: this takes into account the time required by the person to talk, the time required by the cloud server to reply, and the one needed by the robot to speak. 


\begin{remark}
\label{remarktests}
As the system is still in the development stage and it is not yet available to the broader audience, we chose not to perform experiments keeping the server running for long periods, typically to measure the uptime (number of hours the system is running), the availability (the likelihood that a system is available at a specific time instance), and the error rate (the percentage of failed requests out of the total number of requests received by the application). However, our recent experiments report no system failure during $1,125$ minutes of testing with middle school students, a promising indicator of the system's robustness to failures.
\end{remark}

\subsection{Materials and methods}
In the Literature, there are various performance tests for cloud systems \cite{de2017}:
\begin{itemize}
    \item Baseline test: examines how a system performs under expected or normal load and creates a baseline with which other tests can be compared. It aims to find metrics for system performance under normal load;
    \item Scalability test: checks whether the system scales appropriately to the changing load;
    \item Load test: finds metrics for system performance under high load;
    \item Stress test: finds the load volume where the system breaks or is close to breaking;
    \item Soak test: finds system instabilities that occur over time and ensures no unwanted behavior emerges over a long period.
\end{itemize}

Typically, the main factors that can affect the performance of a web service are the payloads (request and response bodies) and the number of requests within a certain amount of time. Therefore, we first focused on executing the Baseline test to evaluate the performance of our system in terms of average response time (i.e., the difference between the time when the request was sent by the client and the time when the response was fully received by the client). We performed this test considering four different payloads (each containing the \textit{client sentence} and a \textit{dialogue state} of a different size) to understand the impact of the request data size on the response time. 

Then, we executed the Scalability test to assess how the average response time increases with a growing number of requests. This aspect is fundamental because it ensures the system is sustainable (regarding response speed), even when many devices use it. The other performance tests were not performed as our current aim is not to assess how the system behaves under high loads or find instabilities over a long activity period, as already clarified in Remark \ref{remarktests}.

\subsubsection{Tools for performance testing}
The web services composing the CAIR server were located on a machine hosted on a cloud service. The machine, running Ubuntu 20.04, was equipped with $16$GB of RAM, 2vCPUs of an Intel Cascade Lake $8260$ @$2.4$Ghz. The clients, simulating multiple robots accessing cloud services, were running on an M1 MacBook Pro with 16GB of RAM, connected to the Wi-Fi network of our laboratory with a Download and Upload speed of approximately $80$Mbps, and a $10$~ms ping. The experiments were performed using Wi-Fi as we assume that most devices would use this type of connection rather than an Ethernet cable. To ensure that the results were not biased by any network overload, we conducted the same experiments using the Ethernet and verified that the response speed difference was always negligible.

The HTTP requests clients make to the server were simulated with the Apache JMeter application. JMeter \cite{halili2008} is a Java application designed to test functional behavior and measure performance. 
Being open-source, simple to use, platform-independent, and suitable to perform any performance test, JMeter is frequently mentioned as a viable solution in the Literature for testing cloud systems \cite{jha2017, wang2019}.
JMeter allows simulating a group of threads (i.e., different clients/robots), sending different requests to a target cloud server. A parameter called ``ramp-up period" tells JMeter how long it takes to ``ramp up" to the full number of threads chosen. If $10$ threads are used, and the ramp-up period is $100$ seconds, then JMeter will take $100$ seconds to get all $10$ threads up and running: each thread will start $10$ seconds (i.e., $100$ seconds divided by $10$ threads) after the previous thread was begun. Once the simulation ends, JMeter returns statistics about the performance of the cloud server. 

For each thread group, in addition to measuring the average response time (provided by JMeter), we also kept track of the average server processing time: that is, the time the Hub takes to respond to the client, measured from the first to the last instruction executed by the server (i.e., without considering the time required by the server for the management of multiple requests on the queue, the creation of threads to handle incoming requests, etc.). The difference between the average response time and the average server processing time provides fundamental information on how the network influences the response time and how contemporary requests are managed on the server side. 

The number of clients concurrently accessing the web services is not the only element that may affect the performance. As already described in Section \ref{sec:client}, the \textit{dialogue state} contains fundamental information about the conversation with a specific client (see Figure \ref{fig:dialogue_state}), and the amount of information encoded in the state grows as the dialogue proceeds. Among the elements that contribute the most to the increase of the \textit{dialogue state}, every time the user expresses a preference about a topic, the corresponding likeliness is modified, and its value is stored in the \textit{dialogue state}. Also, every time a new topic is mentioned, the \textit{dialogue state} must keep a record of the sentences that have already been used by the system when talking about that topic.

\subsubsection{Baseline test}
Given that the \textit{dialogue state} can grow in subsequent iterations, we tested how its increasing size affects the average response time by performing the Baseline test in four scenarios.

\begin{enumerate}
    \item The client has just started interacting with the system. No likeliness value has yet been modified, and no sentence has yet been used by the system. In our case, the payload size was $369$ bytes;
    \item The client has expressed a preference about $1/3$ of the topics present in the Dialogue Tree (i.e., $926$ topics), and information about the sentences associated with those topics has been encoded in the \textit{dialogue state}, increasing the payload size to $6,384$ bytes;
    \item The conversation has gone further, and $2/3$ of the topics have been covered (i.e., $1,853$ topics and $12,213$ bytes);
    \item The client has expressed a preference for all the topics present in the Dialogue Tree (i.e., $2,780$ topics), and information about all the sentences present in the Ontology has been encoded in the \textit{dialogue state}, reaching the maximum payload size of $18,166$ bytes.
\end{enumerate}

For each of the previous scenarios, $30$ requests spaced five seconds apart were performed by a single thread/client. We recorded the response times and the server processing times of each request and then computed their averages.

\subsubsection{Scalability test}
The Scalability test was carried out in two scenarios. 
\begin{enumerate}
\item We simulated an increasing number $N$ of clients/robots performing requests simultaneously;
\item We simulated an increasing number $N$ of clients/robots performing evenly distributed requests over a $10$ seconds period. 
\end{enumerate}

For the first Scalability test, the aforementioned ramp-up period was set to zero, meaning that JMeter started all the threads almost at the same time, while for the second Scalability test, the ramp-up period was set to $10$ seconds. To cover the worst-case scenario, both tests were performed using the greatest request payload in the \textit{dialogue state} (i.e., simulating clients that have expressed a preference for all the topics in the DT). For each scenario, we measured the response times and server processing times of the group of $N$ threads for $30$ times, with the $N$ threads distributed over the respective ramp-up period. Each group of $N$ threads started approximately five seconds after the previous group had ended. Then, we computed the average response time and processing time versus the number $N$ of users in the two configurations (i.e., simultaneously or over $10$ seconds).

\subsubsection{Tests on the field}
Tests on the field during actual human-robot interaction were performed, by measuring the average response time during our recent tests on the field with $300$ high-school students, providing insight into the system's usability in real-world scenarios. Specifically, we compared response times using a 4G connection to access the Cloud and a Wi-Fi connection to a local server. 

\subsubsection{Acceptable response time}
For these experiments, we established a threshold of one second as the maximum acceptable average response time. Please notice that the threshold is below the delay reported in experiments about people's perception during a dialogue with a conversational system \cite{peng2020, shiwa2008}. Setting the threshold to one second arranges for variations due to the load, the network performance, or the additional time required to perform the speech-to-text transcription. This ensures higher satisfaction during the conversation.

\subsection{Results}
\label{sec:results}
\begin{figure}[ht]
    \centering
    \includegraphics[width=\linewidth]{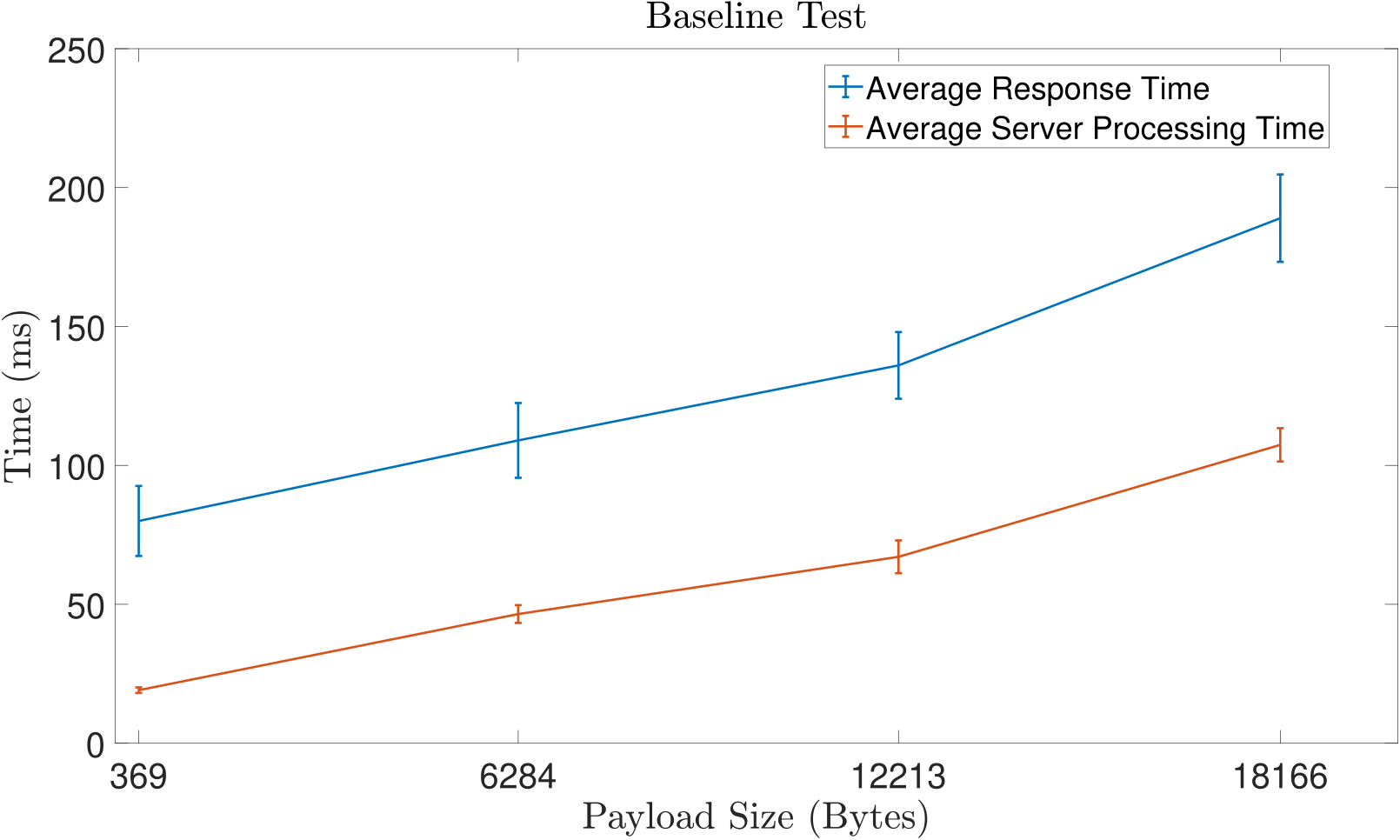}
    \caption{Results of the Baseline test with the 4 payload sizes.}
    \label{fig:baseline_test}
\end{figure}

Figure \ref{fig:baseline_test} shows the results of the Baseline test. On the x-axis, we reported the payload size of the request in the four considered scenarios. The y-axis shows the time in ms, computed as the average of $30$ independent requests (i.e., with no overlapping time). The blue line represents the average response time, while the orange line depicts the average server processing time.

\begin{figure}[t]
    \centering
    \includegraphics[width=\linewidth]{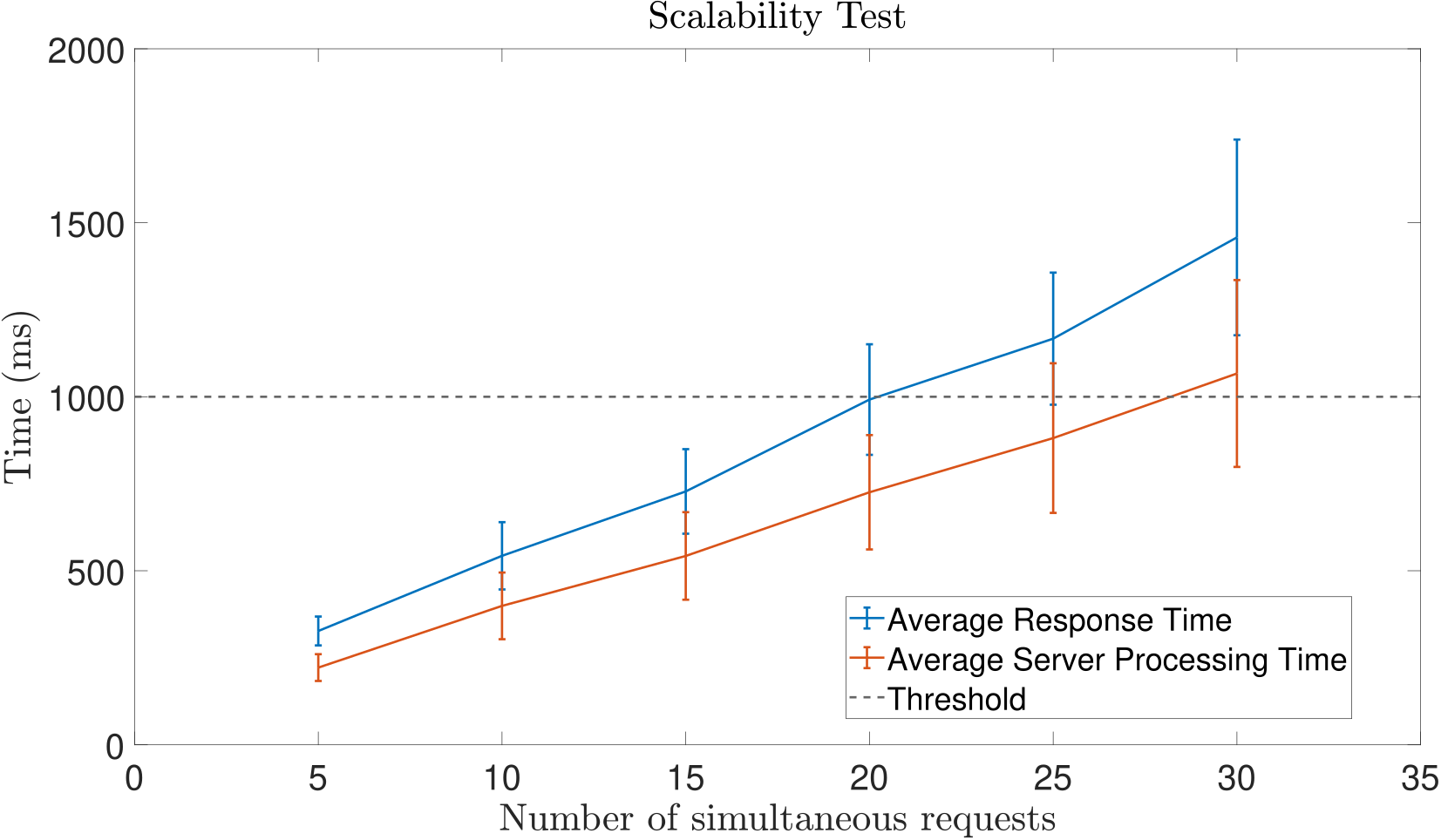}
    \caption{Results of the Scalability test with a growing number of simultaneous requests and a threshold of one second.}
    \label{fig:scalability_test_0}
\end{figure}

\begin{figure}[t]
    \centering
    \includegraphics[width=\linewidth]{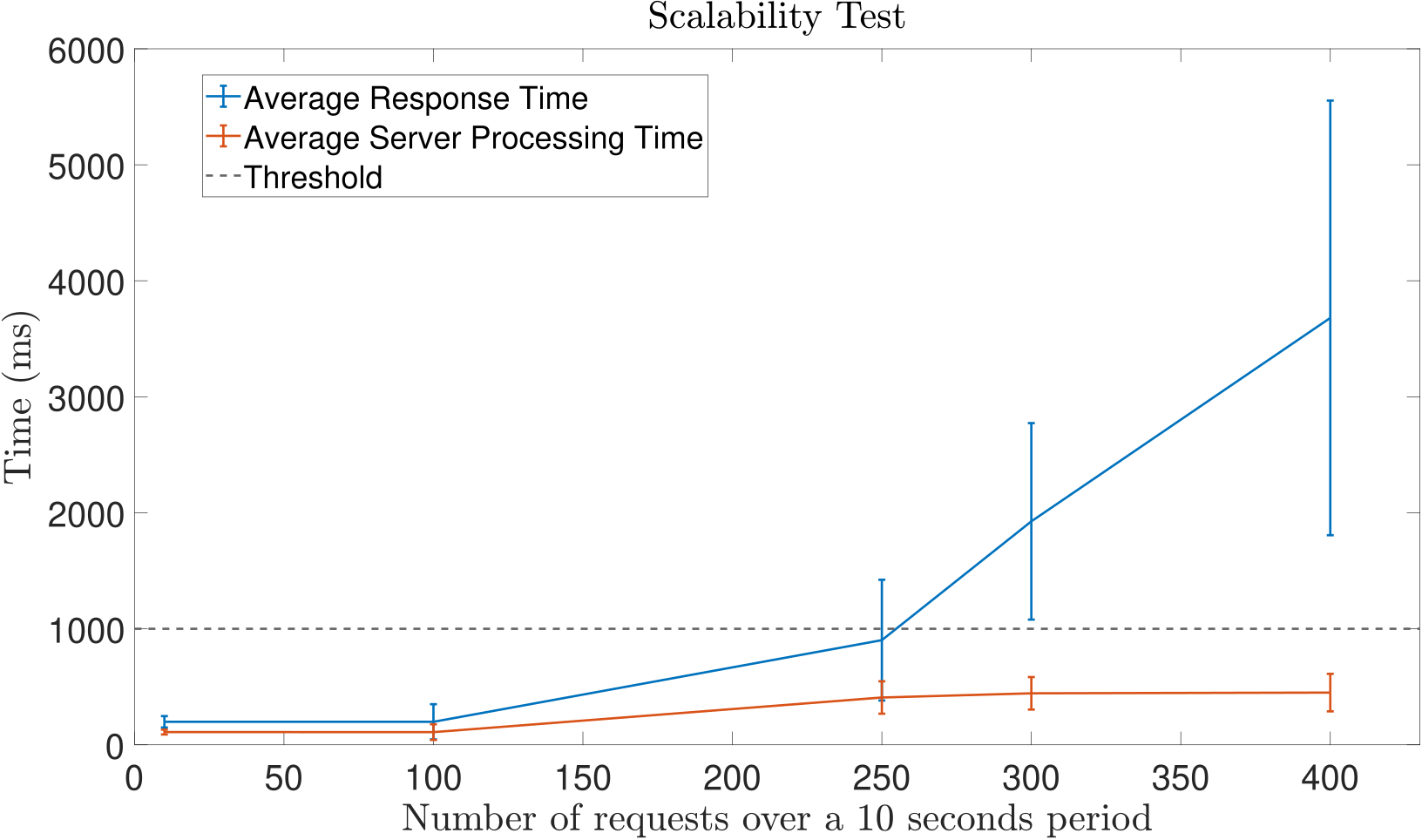}
    \caption{Results of the Scalability test with a growing number of requests performed in a $10$ seconds period and a threshold of one second.}
    \label{fig:scalability_test_10}
\end{figure}

Figure \ref{fig:scalability_test_0} depicts the results of the Scalability test carried out with $N$ contemporary requests performed by $N$ threads, while Figure \ref{fig:scalability_test_10} reports the results of the Scalability test performed distributing the $N$ requests over a $10$ seconds period. In both cases, the x-axis reports the number $N$ of requests performed during the respective ramp-up period, while the y-axis shows the time in ms, computed as the average of $30$ independent tests (each with $N$ users). As in the previous graph, the blue line reports the results for the average response time, while the orange line depicts the results for the average server processing time. 

In all graphs, the dotted line highlights the one-second threshold for the acceptable response time. 

We confirmed the results of our tests with field experiments with middle school students. The tests yielded an average response time of $0.26~s$ (standard deviation $0.06$) when using a 4G connection to the cloud server and $0.17~s$ (standard deviation $0.08$) when connecting to a local server through Wi-Fi. It is important to note that 4G response times depended on the available bandwidth during each experiment, measured through Ookla speedtest. Additionally, please note that the time required to convert audio samples into text (using Microsoft Azure API) was not included in this computation.

\begin{remark}
Throughout all of our experiments, the bandwidth usage (measured with a speedtest\footnote{\url{https://www.speedtest.net}}) was consistently below the maximum capacity of the connection used.
\end{remark}

\subsection{Discussion}
\label{sec:discussion}
By looking at Figure \ref{fig:baseline_test} (Baseline test), it can be noticed that the standard deviations of the response times are always higher than the processing times: this may be due to fluctuations in the speed of the network. Also, when performing the Baseline test with the greatest payload, the average response time reached a value of $189$~ms, while the average server processing time was $107.4$~ms. This difference is mainly due to the network latency that strongly influences the response time. However, this test suggests that, even in the worst-case scenario (i.e., with the largest payload size), the performance of the system under a normal load (i.e., non-overlapping requests) guarantees a high level of satisfaction of the user interacting with the system since values are below the acceptability threshold.

From Figure \ref{fig:scalability_test_0} (Scalability test), it is immediate to notice that, as in the previous test, the standard deviations of average response times are always higher than those of average server processing time. When considering the average response time, the one-second acceptability threshold is reached with approximately $20$ simultaneous requests, while the average server processing time reaches the acceptability threshold with around $50$ contemporary requests. Measuring the server processing time shall be interpreted as a limit situation to measure how the system will behave in case no network delay is present (simulating the case of a local server). 

Observing the results in Figure \ref{fig:scalability_test_10} (Scalability test), in addition to the huge difference between the standard deviations, it can be seen how the average response time reaches the threshold with around $250$ requests distributed over a $10$ seconds period, while the average server processing time tends to stabilize at around $450$~ms. This is because the server can manage up to a certain number of contemporary requests, and those arriving after are queued and have to wait for some time before being processed. The server processing time, measured from the first to the last instruction executed by the server, tends to be constant. Delays are mostly due to the network and the queuing mechanism for managing a huge number of concurrent requests.

Given these results, it is interesting to make some considerations about the maximum number of users that the server can handle. We assumed that, when interacting with the system, a user can perform, on average, six requests per minute. However, it is reasonable to assume that among all users $M$ that subscribed to the system and can virtually use it, only a ratio $R$ will concurrently access its services during a given period of the day. This yields $N=RM$ users that can concurrently make a request. 

Computing $R$ is not the purpose of this work. Its computation should be based on several considerations, including users' geographical distribution, the times of the day when they are more likely to connect to the system, how long they use the system on average, and so on. Suppose that $R=0.2$: if $200$ users are subscribed to the cloud services, it turns out that $N=40$ users may concurrently connect to the services. By reasoning in the opposite way, knowing $R$ plays a key role in cloud sizing as it allows us to know the number of users $M$ who can subscribe to the system, depending on the number of users $N$ that can concurrently connect to the cloud without compromising the quality of service. That is, if we know the maximum number $N$ of concurrent users (in our tests, $N=20$ in the unrealistic case of simultaneous requests, and $N=250$ in case the request are uniformly distributed), we can estimate the maximum number of users that can subscribe to the system, as $M=N/R$. With $R=0.2$, $M=100$ if we want to ensure a 1-second delay under any possible condition, while $M=1250$ in case we aim to average performance.

As a final consideration, it may be observed that the proposed approach, despite the benefits originating from not storing the conversation state on the server, has obvious limitations. If the Ontology is expanded and more conversation topics are added, the maximum size of the \textit{dialogue state} grows, leading to an increase in the response time. Our previous and ongoing experiments with care home residents \cite{papadopoulos2020, Papadopoulos2021}, high-school students, and patients with spinal cord injuries revealed that the current Ontology with about $3,000$ topics of conversation is quite comprehensive to allow users to have an engaging conversation with the system. Experiments also revealed that during a conversation, it is quite unlikely that the situation in which all $3,000$ topics have been explored occurs. This situation would produce a \textit{dialogue state} that reaches the maximum theoretical upper bound. However, people tend to repeatedly explore topics they like the most (and this is especially true for older people). The system allows the users to express their own opinion and tell their own stories by paying interest in what they say, a possibility that people appreciate more than listening to what the robot has to say. If these considerations are not confirmed in the future, moving part of the \textit{dialogue state} to the cloud is a possibility to be explored.

\section{Conclusion}
\label{sec:conclusion}
The work presented the architecture of a cloud system designed to allow for interaction with social robots and other conversational agents. Our proposal is based on the consideration that a huge number of low-cost smart devices for social interaction are expected to hit the market soon and will not have the onboard computational capabilities to perform the complex operations required for interacting with humans. Under these conditions, we need a system that is sustainable in three different ways.
\begin{itemize}
    \item It should provide a scalable solution with acceptable performance and cost, which can be easily expanded by adding new services to improve the capabilities of the robots and devices connected to the cloud. 
    \item It should be usable by any robots and devices with Internet connectivity, able to acquire an input through a keyboard or a microphone and provide an output through a screen or a speaker by developing a proper client that connects to its services. 
    \item It should be fully customizable but also provide a complete solution for a conversation about several topics that researchers and companies can use without requiring additional programming or customization.
\end{itemize}

The server includes an easily expandable portfolio of web services that comply with REST rules. A Hub service collects the requests and redirects them to the Plan Manager and the Dialogue Manager services. The former is in charge of recognizing the user's intention to execute an action, while the latter manages the dialogue. To provide appropriate plans and answers, these services exploit the information encoded into an Ontology, specially designed to take into account the cultural background of the user. The cultural aspect of the Ontology is not discussed in detail in this article (see \cite{bruno2019, recchiuto2020}).


The experiments carried out aimed at assessing the performance of the system in terms of response speed. The Baseline test was meant to investigate how the system performed under a typical load. The results of the test showed that even with the maximum payload, the average response time is within $200$~ms. The Scalability tests had the objective of investigating whether the system scaled appropriately to an increasing load. The first one revealed that the system could support up to $20$ simultaneous requests, while the second proved that about $250$ users could perform evenly distributed requests over $10$ seconds without exceeding the acceptability threshold of one second on the average response time. 

These findings will give us the basis to size the system, paving the way to a sustainable solution for verbal interaction with low-cost robots and other smart devices.

\backmatter

\section*{Declarations}
\begin{itemize}
\item Funding: not applicable.
\item Conflict of interest: the authors have no conflicts of interest to declare. All co-authors have seen and agree with the contents of the manuscript and there is no financial interest to report. We certify that the submission is original work and is not under review at any other publication.
\item Ethics approval: not applicable.
\item Availability of data and materials: data and materials are available on request.
\item Code availability: all the developed clients are available on request. 
\end{itemize}

\bibliographystyle{unsrt}
\bibliography{main}

\end{document}